\documentclass[conference]{IEEEtran}
\IEEEoverridecommandlockouts
\usepackage{cite}
\usepackage{multirow}
\usepackage{subcaption}
\usepackage{amsmath,amssymb,amsfonts}
\usepackage{algorithmic}
\usepackage{graphicx}
\usepackage{textcomp}
\usepackage{xcolor}
\usepackage{caption}
\usepackage{hhline}
\usepackage{arydshln}
\usepackage{color}
\usepackage{tabularray}
\usepackage{dirtytalk}
\usepackage{siunitx}
\usepackage{array}
\usepackage{pifont}
\usepackage{colortbl}
\usepackage{verbatim}
\usepackage[font=small]{caption}

\definecolor{Silver}{rgb}{0.752,0.752,0.752}
\definecolor{SurfCrest}{rgb}{0.815,0.901,0.815}
\def\BibTeX{{\rm B\kern-.05em{\sc i\kern-.025em b}\kern-.08em
    T\kern-.1667em\lower.7ex\hbox{E}\kern-.125emX}}
\begin{document}

\title{%AIR-TPC: Artificial Intelligence Radio for real-time Transformer-based Protocol Classifier  
T-PRIME: \underline{T}ransformer-based \underline{Pr}otocol \underline{I}dentification for \underline{M}achine-learning at the \underline{E}dge\\

\thanks{\tiny © 2024 IEEE. Personal use of this material is permitted. Permission from IEEE must be obtained for all other uses, in any current or future media, including reprinting/republishing this material for advertising or promotional purposes, creating new collective works, for resale or redistribution to servers or lists, or reuse of any copyrighted component of this work in other works.}
}

\author{
\IEEEauthorblockN{Mauro Belgiovine,
Joshua Groen, 
Miquel Sirera,
Chinenye Tassie,
Ayberk Yark{\i}n Y{\i}ld{\i}z,
Sage Trudeau, \\
Stratis Ioannidis, %\IEEEauthorrefmark{2}, 
Kaushik Chowdhury } %, \IEEEauthorrefmark{2}}
\IEEEauthorblockA{Department of Electrical and Computer Engineering,
Northeastern University
Boston, MA \\
Email:\{belgiovine.m, groen.j, sirera.m, tassie.c, yildiz.ay, trudeau.s, ioannidis, k.chowdhury\}@northeastern.edu}
}

\maketitle

% %IEEE Copyright
% % \begin{picture}(0,0)(15,-190)
% \begin{picture}(0,0)(15,550)
% \put(0,0){
% \footnotesize © 2024 IEEE. Personal use of this material is permitted. Permission from IEEE must be obtained for all other uses, in any current or future media, including}
% \put(0,-10){
% \footnotesize reprinting/republishing this material for advertising or promotional purposes, creating new collective works, for resale or redistribution to servers or lists, or}
% \put(0,-20){
% \footnotesize reuse of any copyrighted component of this work in other works.}
% \end{picture}

% © 2024 IEEE. Personal use of this material is permitted. Permission from IEEE must be obtained for all other uses, in any current or future media, including reprinting/republishing this material for advertising or promotional purposes, creating new collective works, for resale or redistribution to servers or lists, or reuse of any copyrighted component of this work in other works.

\begin{abstract}
Spectrum sharing allows different protocols of the same standard (e.g., 802.11 family) or different standards (e.g., LTE and DVB) to coexist in overlapping frequency bands. As this paradigm continues to spread, wireless systems must also evolve to identify active transmitters and unauthorized waveforms in real time under intentional distortion of preambles, extremely low signal-to-noise ratios and challenging channel conditions. We overcome limitations of correlation-based preamble matching methods in such conditions through the design of T-PRIME: a Transformer-based machine learning approach. T-PRIME learns the structural design of transmitted frames through its attention mechanism, looking at sequence patterns that go beyond the preamble alone. The paper makes three contributions: First, it compares Transformer models and demonstrates their superiority over traditional methods and state-of-the-art neural networks. Second, it rigorously analyzes T-PRIME's real-time feasibility on DeepWave's AIR-T platform. Third, it utilizes an extensive 66 GB dataset of over-the-air (OTA) WiFi transmissions for training, which is released along with the code for community use. Results reveal nearly perfect (i.e. $>98\%$) classification accuracy under simulated scenarios, showing $100\%$ detection improvement over legacy methods in low SNR ranges, $97\%$ classification accuracy for OTA single-protocol transmissions and up to $75\%$ double-protocol classification accuracy in interference scenarios.

%Older version
%As new frequency bands are released for commercial applications, the demarcation between licensed and unlicensed spectrum is becoming increasingly blurred. While this wider selection of frequencies offers potential solutions, the escalating number and diversity of wireless devices exacerbate the issue of spectrum scarcity. Consequently, the need for effective multi-protocol spectrum sharing methods becomes paramount. With their self-attention mechanism, transformers neural networks capture global dependencies and long-range interactions, enabling effective modeling of complex relationships within wireless signals. This capability makes transformers particularly well-suited for protocol classification, as they can capture the intricate patterns and characteristics that differentiate between various wireless protocols, even in the presence of noise, interference, and channel distortions. In this work we present first a comparative simulation study on the performance of transformer versus convolutional neural network (CNN) architectures for WiFi protocol classification under different channel models and noise levels. Once  transformer superiority over CNN is established, we deploy the proposed model on DeepWave's Artificial Intelligence Radio Transceiver (AIR-T) and test it with real-world over-the-air (OTA) transmissions. Our approach demonstrates a $> XX\%$ classification accuracy under simulated and OTA scenarios and an increase of $YY\%$ detection accuracy over traditional methods.
\end{abstract}


\section{Introduction}
%The proliferation and the ever-increasing demand for wireless services have led to the scarcity of available spectrum resources \cite{zhang2022machine}. As a result, various wireless communication protocols coexist within the same frequency bands, leading to congested wireless spectrum environments \cite{schmidt2017wireless}. Moreover, security concerns arise when unauthorized transmissions occupy or interfere with unintended frequencies, posing risks to first responders and other critical operations \cite{jagannath2021multi}. %In such scenarios, effective protocol classification emerges as a vital mechanism for efficient spectrum management and utilization. Protocol classification entails the identification and categorization of wireless communication protocols operating within a given frequency range. 
%By detecting the diverse protocols present in a crowded spectrum, regulatory bodies, network operators, and researchers can devise intelligent strategies to mitigate interference, improve spectral efficiency, and enhance overall wireless system performance. 

The increasing demand for wireless services has caused a scarcity of spectrum resources \cite{zhang2022machine}. This results in congested wireless spectrum environments as various communication protocols coexist in the same frequency bands \cite{schmidt2017wireless}. Unauthorized transmissions further raise security concerns, posing risks to critical operations \cite{jagannath2021multi}. Detecting diverse protocols in crowded spectrums allows for intelligent strategies to mitigate interference, improve spectral efficiency, and enhance overall wireless system performance, benefiting regulatory bodies, network operators, and researchers.

\begin{figure}[t]
    \centering
    \includegraphics[width=9cm]{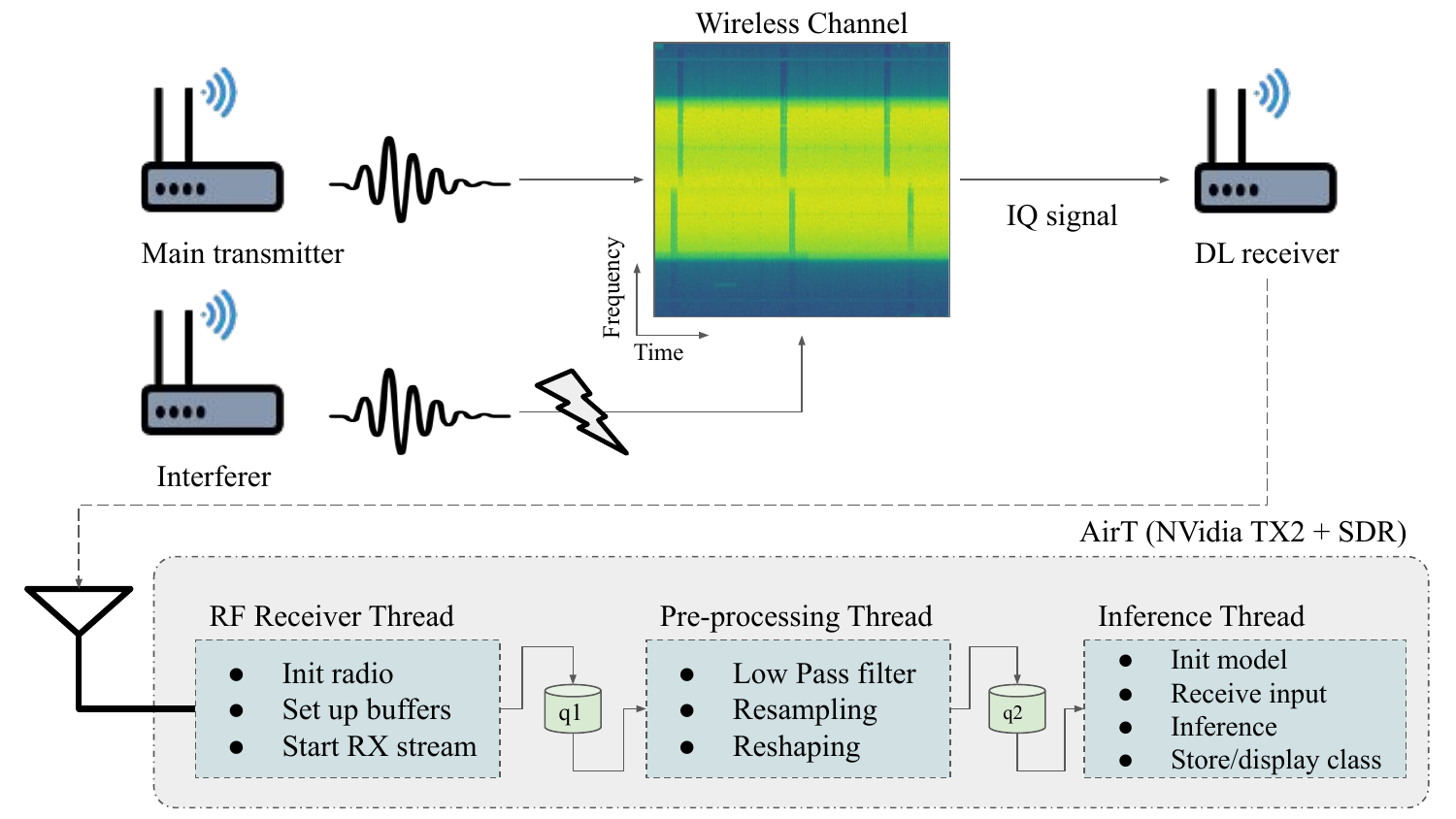}
    \caption{T-PRIME system overview and receiver design. }
    \label{Fig: system overview}
\end{figure}

\noindent $\bullet$ \textbf{Problems with legacy methods of protocol classification.} %
Legacy methods require extensive prior knowledge of protocols, leading to challenges in detecting new protocols. %in dynamic spectrum sharing scenarios. 
Even with full protocol knowledge, correlation-based methods suffer from reduced detection accuracy in low SNR conditions and when multiple protocols overlap in frequency. To address these challenges, T-PRIME (shown in Fig. \ref{Fig: system overview}) adopts a Machine Learning (ML) based approach, performing protocol classification without requiring specific channel boundaries or symbol rates, even with overlapping signals and low SNR conditions.

Legacy protocol detection methods are integrated into the RF receiver network interface card, enabling fast preamble correlations in hardware. However, updating the system for new protocols leads to backward compatibility issues and potential detection errors in challenging wireless channels. On the other hand, software-defined radio (SDR) systems offer flexibility but introduce higher latencies due to data transfer and complexities in I/O and buffer management. This paper demonstrates that using edge devices with CPU and GPU on a system-on-a-module (SOM), along with careful software design, can overcome these limitations and enable real-time processing for complex ML wireless applications on an SDR-based edge device.

\noindent $\bullet$ \textbf{Problems with WiFi protocol classification.} We showcase the potential of SDR based, ML enabled architectures by focusing on a challenging wireless scenario involving protocol classification. Specifically, we classify different 802.11 WiFi standards, including 802.11b, 802.11g, 802.11n, and 802.11ax. These  standards share similar preamble structures and can operate with the same MCS, frequency bands, channel bandwidths, and transmit power. Importantly, this classification is distinct from and more difficult than prior works \cite{zhang2022machine, schmidt2017wireless, jagannath2021multi}, which classify diverse protocols like LTE, WiFi, and Bluetooth from a candidate pool and lack a real time deployment.

\noindent $\bullet$ \textbf{Proposed solution: T-PRIME.} 
% Transformers are the base of LLMs such as ChatGPT etc..
We demonstrate the effective use of Transformer neural networks \cite{NIPS2017_3f5ee243} for the above WiFi protocol classification problem. Transformers have gained significant interest in the deep learning community for their ability to ingest long sequences of data and efficiently correlate temporal features far from each other, which is particularly useful for natural language processing tasks \cite{wolf-etal-2020-transformers}. However, their application to wireless communications is still in a nascent stage, with some works focusing on modulation classification \cite{9779340, zheng2022fine, hamidi2021mcformer}.
%
%Although Transformer-based models excel at processing sequential and time-dependent information, adapting them to work with IQ samples in the RF domain poses challenges. 
Although Transformer-based models excel at processing 1D sequential and time-dependent information, adapting them to work with 2D sequences of IQ samples is not straightforward.
Additionally, it is crucial to evaluate ML models intended to be deployed as real-time classifiers on resource-constrained devices to understand the trade-offs between accuracy and delay.
Our contributions are as follows:
%We demonstrate how Transformer neural networks can be effectively used for the above problem of WiFi protocol classification. The Transformer architectures with attention mechanism \cite{NIPS2017_3f5ee243} have seen significant interest recently within the deep learning community. They demonstrate remarkable proficiency in natural language processing tasks~\cite{wolf-etal-2020-transformers}, although their use for wireless communications remains at a nascent stage with early works related to modulation classification ~\cite{9779340, zheng2022fine, hamidi2021mcformer}. 
%to accurately identify protocols under challenging scenario, such as overlapping interference cases and very low SNR levels (i.e. $\le 0$ dB). 
%While Transformer-based models excel at processing sequential and time-dependent information, helping them to learn the subtle features of wireless signals, adapting them to use IQ samples in the RF domain is not trivial. We must also evaluate ML models deployed as real time classifiers on resource constrained devices to understand the cost benefit tradeoffs of accuracy and delay as a function of model size.

\begin{itemize}
    \item We create, and publicly release \cite{tprime_github}, a protocol classifier called T-PRIME based on Transformer architecture and compare it to both classical, preamble based processing and  ML based modulation detectors using a large data set of simulated WiFi signals. %Specifically, we show how the Transformer model operates with minimal knowledge of the protocols under study, as opposed to legacy methods.  
    %We study the impacts of Transformer model size on real time inference on an edge-device.
    
    \item We rigorously evaluate the performance of each of these protocol classification methods under various Single-Input-Single-Output (SISO) wireless channel settings, including randomized channel realization, different signal-to-noise ratio (SNR) levels, and multiple channel models. In all these settings, T-PRIME outperforms state-of-the-art legacy and ML solutions available for classification of raw IQ signals.
    
    \item We implement T-PRIME on DeepWave's Artificial Intelligence Radio Transceiver (AIR-T) and evaluate performance in realistic scenarios. Drawing from this experience, we provide insights that will allow building a generic  ML-based wireless signal processing pipeline on SOM architectures that operates in near real-time using sequences of time-domain IQ samples.
   
    \item We obtain and release the first-of-its-kind dataset comprising over-the-air (OTA)  WiFi-signals for different protocol versions. This dataset includes transmissions collected in diverse environments, encompassing both single and overlapping transmission scenarios. %(e.g., with an additional interfering signal occupying either $25\%$ or $50\%$ of the intended transmitter's band).
    
\end{itemize}

\section{Related Work}

\noindent $\bullet$ \textbf{Preamble correlation Methods.}
Frame preamble usage for protocol identification and synchronization involves detecting a predefined symbol sequence at the start of a transmitted signal. %, which serves as a unique identifier for a wireless protocol. 
For example, the 802.11b long preamble consists of 128 scrambled 1s and 16 SFD marker bits, while the 802.11n preamble includes at least 9 symbols spread across all 52 OFDM sub-carriers.

Traditional signal processing techniques use energy detection and correlation \cite{9363693, terry2002ofdm}, requiring prior knowledge of the unique preamble sequences. However, they can be vulnerable to noise, interference, and channel variations. In Sec. \ref{sec:legacy_performance_simulation}, we demonstrate %through extensive simulations 
how legacy approaches fail at low SNR, while T-PRIME achieves highly accurate protocol classification under the same conditions.

%The use of the frame preamble for protocol identification and synchronization involves detecting a predefined sequence of symbols at the beginning of a transmitted signal. The  preamble serves as a unique identifier for a particular wireless protocol. For example, the 802.11b long preamble is a pattern of 128 scrambled 1s plus 16 SFD marker bits while the 802.11n preamble has at least 9 symbols spread over all 52 OFDM sub-carriers. Traditional signal processing techniques use a combination of energy detection and correlation~\cite{9363693}, which requires prior knowledge of the unique preamble sequences. %However, they often %require prior knowledge of the expected preamble sequences or templates for different protocols.  
%Moreover, they may be susceptible to noise, interference, and variations in channel conditions. Indeed, we demonstrate in Sec. \ref{sec:legacy_performance_simulation} through extensive simulations how legacy approaches fail at low SNR, while T-PRIME is able to correctly classify protocols with extremely high accuracy under the same conditions.
%Advanced techniques, including machine learning and deep learning approaches, are increasingly being employed to enhance preamble detection by leveraging the capabilities of these traditional signal processing techniques while improving robustness and adaptability in complex and dynamic wireless environments.

\noindent $\bullet$ \textbf{ML-based methods.}\label{background-ml}
Protocol classification via ML methods is still in a nascent stage. Zhang \emph{et al.} ~\cite{zhang2022machine} perform protocol detection between WiFi, LTE, and 5G signals using a multi-layer bidirectional RNN. %They utilize Matlab to generate signals, apply an AWGN channel, and capture the resulting IQ samples. 
They use 512 IQ pairs with a sliding window as inputs to their RNN. %However, they do not distinguish between different WiFi protocols. 
Similarly, Schmidt \emph{et al.} use a CNN to classify between WiFi, Bluetooth, and ZigBee signals \cite{schmidt2017wireless}. Jagannath and Jagannath  \cite{jagannath2021multi} perform both modulation classification and signal classification via CNNs by distinguishing between a wide range of different protocols including AM Radio, Bluetooth, WiFi, and IEEE 802.15.4  (Zigbee). %They use a multi-task learning model that has a shared convolutional and max-pool layer as well as task specific convolutional and fully connected layers. 
However, these approaches do not address the challenge of classifying similar protocols, such as different WiFi standards. %, making T-PRIME the first real-time 802.11 WiFi protocol classification system.

%Protocol classification is more challenging when the candidate protocols are similar in terms of modulation classes and frame structure, such as between different WiFi standards, which is not explored in prior work. Thus, T-PRIME addresses a key gap in the state-of-the-art as the first 802.11 WiFi protocol real-time classification system. 

There is a growing body of work that uses ML for the different problem of modulation classification.  These include CNN-based architectures \cite{o2016convolutional, shi2019deep, elyousseph2021deep, 8267032, gravelle2019sdr, huynh2020mcnet}, CNNs with a self-attention mechanism \cite{zhang2023amc} and RNNs \cite{8357902}. There is a smaller number of prior works that introduce Transformer based ML models for the modulation classification task. For example, Zhen \emph{et al.} create  spectrograms from IQ data and use an image Transformer \cite{zheng2022fine}. Hamidi-Rad and Jain \cite{hamidi2021mcformer} first use a CNN to transform the synthetically generated IQ samples to a 1 dimensional latent space input to the Transformer. They utilize this input CNN layer to help extract features in a location independent way because their IQ samples are not synchronized and start and stop at random locations.  Cai \emph{et al.} \cite{9779340} state that IQ data can not be directly applied to the Transformer because it represents only two sequences (I and Q) and a Transformer requires several sequences. They utilize a four step pre-processing technique that creates \say{patches} of IQ samples that are temporally close to each other. % from synthetic data sets. 
Finally, they use a linear projection embedding layer with additional sequence and positional encoding before the Transformer encoder layer. In contrast to these works that use complicated signal pre-processing and show results on offline inference only, T-PRIME shows how IQ samples can be directly provided as the input to a Transformer and performs well in realistic environments on an edge platform.
%These works are solving a fundamentally different problem than protocol detection. 
%The majority of prior modulation classification works, such as \cite{8977561, shi2019deep, elyousseph2021deep, 8267032, gravelle2019sdr, zhang2023amc, huynh2020mcnet}, use CNN or RNN based ML models for modulation classification. 

%While each of the 802.11 protocols support multiple modulation types, we specifically use 16-QAM for g, n, and ax and QPSK for b. Furthermore, the preamble for all 802.11 phy layer frames is sent using BPSK modulation. Thus, a modulation detector alone cannot distinguish between different 802.11 protocols.

%Prior work has mainly utilized CNN or ResNet-based models.
 %However, they each approach the input to the transformer in different manners. 
 %To the best of our knowledge there are no prior works that use ML to classify different 802.11 WiFi protocols.  

\section{T-PRIME Design}

Transformers, as detailed in \cite{niu2021review}, employ attention mechanisms, or activation masks, which filter information within a layer for transmission to the subsequent layer. In \emph{self-attention}, the network generates this mask based on input and/or preceding layers. Transformers excel in modeling varied dependencies among sequence elements, independent of their positions in input or output sequences, unlike RNNs. Crucially, these adaptable dependencies in Transformers are shaped by and learned from input data. %Finally, Transformers mitigate the \say{dampening} effect found in RNNs across layers. 

\begin{figure}[t]
    \centering
    \includegraphics[width=.9\linewidth]{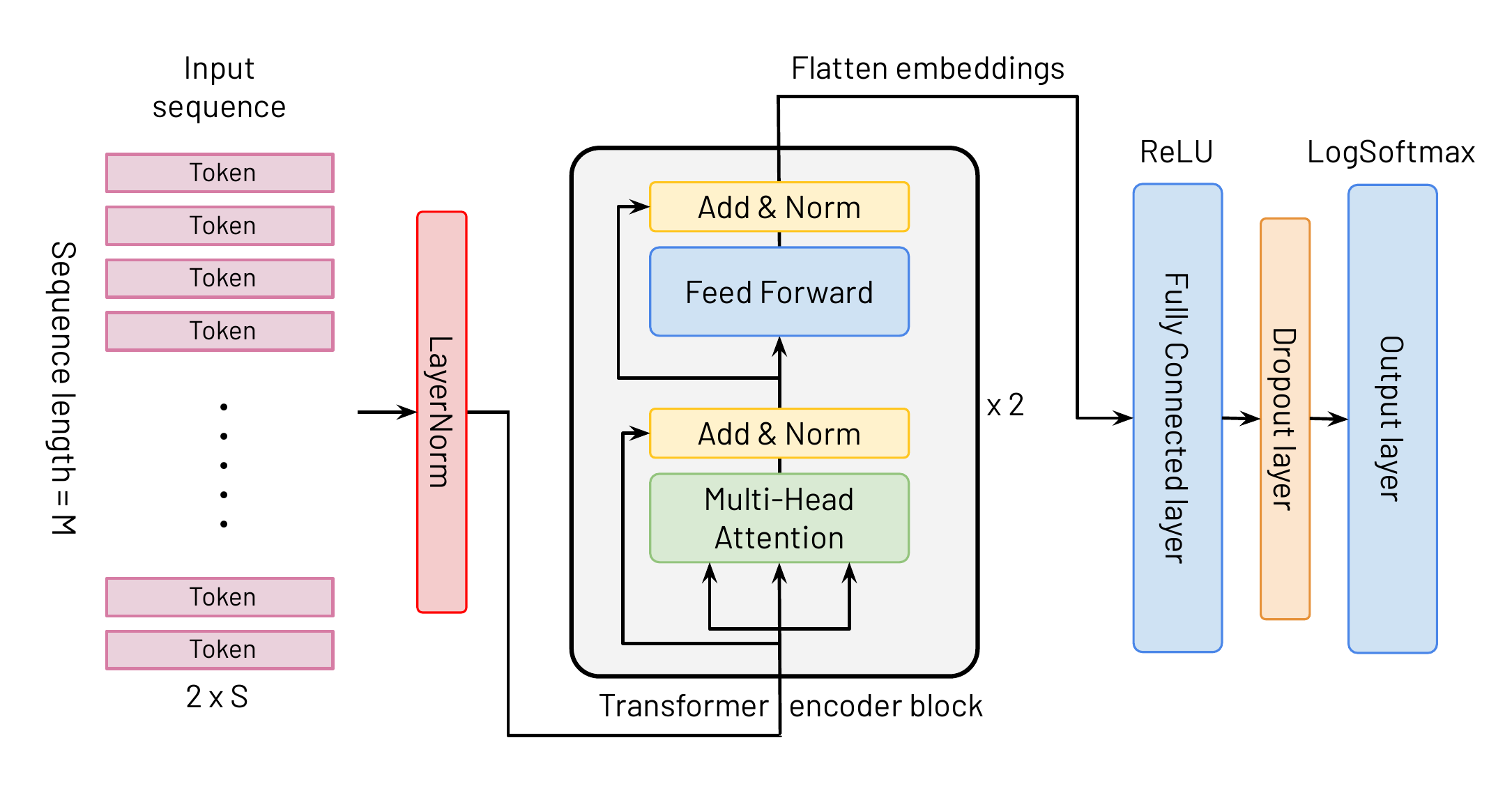}
    \caption{Transformer-based model architecture. %Parameters M, S of the input size will be different for the two model sizes. For the small architecture (SM), M=24, S=64 and for the large one (LG) M=64 and S=128.
    }
    \label{Fig: Transformer arch}
\end{figure}

The general Transformer architecture, as described by Vaswani \textit{et al.} \cite{NIPS2017_3f5ee243}, is composed of several stacked transformer layers shown in Fig.~\ref{Fig: Transformer arch}, typically forming an encoder-decoder structure. Each layer has two sub-layers: a multi-head attention mechanism and a feed-forward network. The input data, along with positional encoding, is fed into the Transformer. Next, it is passed to the multi-head attention mechanism. There are residual connections and normalizations around the two sub-layers. Implementations generally differ in where the normalization occurs and in the specific implementation of the feed-forward layer.

\begin{figure}[t]
    \centering
    \includegraphics[width=.8\linewidth]{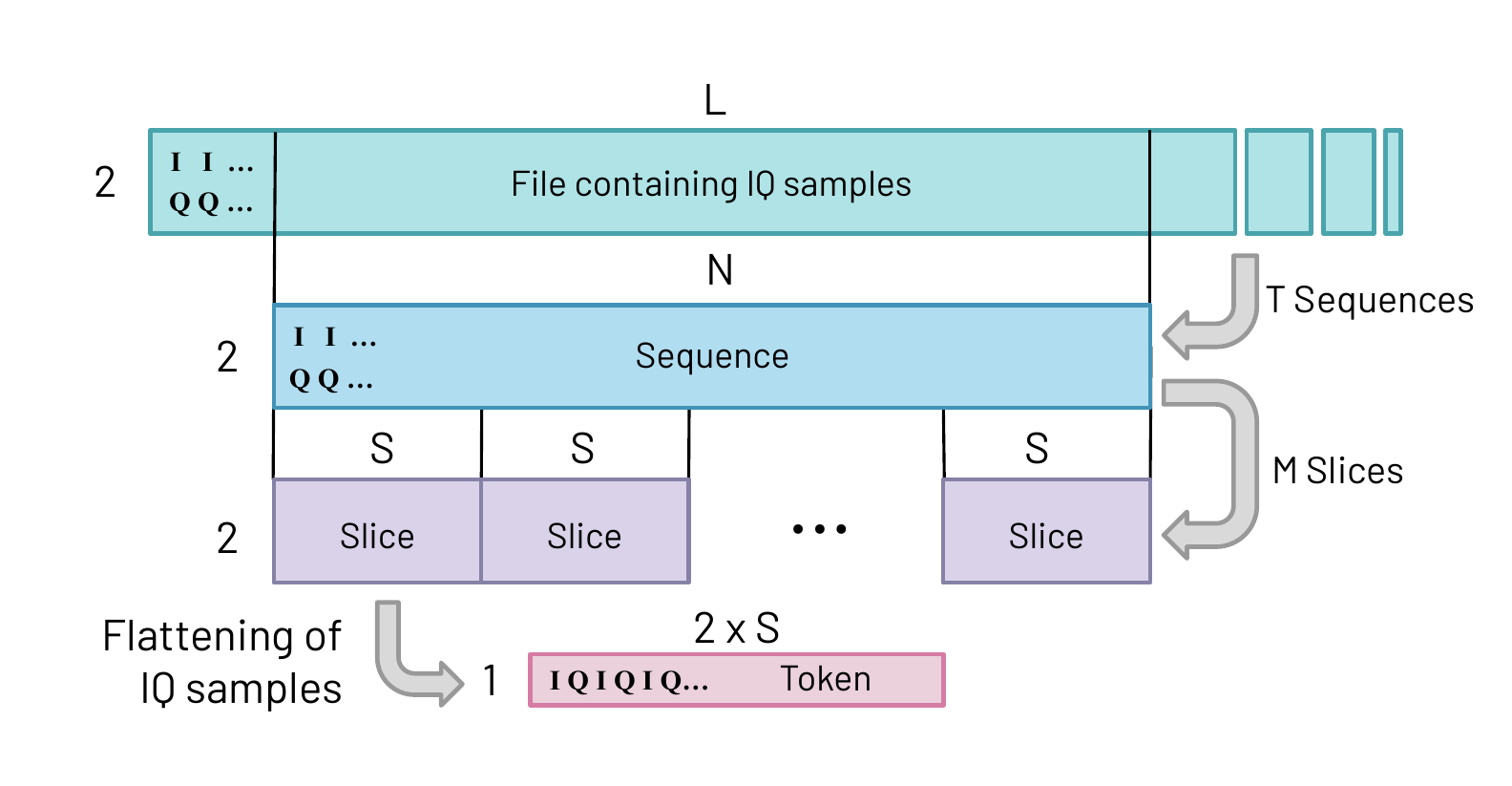}
    \caption{Transmission split into sequences, slices and, tokens.}
    \label{Fig: IQtotokens}
\end{figure}

\noindent $\bullet$ \textbf{Adapting Transformer for IQ Input.} The classical Transformer architecture requires a token sequence as input, typically achieved in NLP tasks through word embeddings and positional information. However, feeding IQ data to the Transformer is not straightforward, leading to complex approaches in prior work \cite{zheng2022fine, hamidi2021mcformer, 9779340}. Initially, we used spectrograms and a vision Transformer \cite{dosovitskiy2020image}, but this yielded lower performance compared to a baseline CNN \cite{o2016convolutional}.

To address this, we adopted an NLP-inspired approach of breaking the IQ samples into smaller \say{sentences} and \say{words}. Our dataset initially consists of $2 \times L$ samples, which are partitioned into sequences of dimension $2 \times N$ (sentences), then further sliced into $M$ slices of size $2 \times S$ (words). Instead of defining a linear embedding for these \say{words} as in traditional NLP Transformers %we recognized that they already represent a high-dimensional embedding space. 
and since I and Q real values are defined in the [0,1] interval, we treat the $2S$-dimensional space defined by the set of all possible \say{words} as our embedding space.
Thus, we simply flattened them by interleaving the I and Q samples into a vector of dimension $1 \times 2S$, eliminating the need for linear embeddings (see Fig.~\ref{Fig: IQtotokens}).

\noindent $\bullet$ \textbf{Transformer Architecture Design.} 
Our proposed Transformer architecture (Fig. \ref{Fig: Transformer arch}) omits positional encoding as it did not affect accuracy. Since our task is classification, we utilize only the encoder part of the Transformer. Contextualized embeddings from the encoder layers flow into a Fully-Connected layer and an Output layer representing the four classes in our classification. The architecture includes an initial normalization layer, \textit{LogSoftmax} as the final activation function and \textit{NLLLoss} as loss function.

We evaluate T-PRIME using a systematic two-phase approach. In the first phase, we generate a large synthetic dataset to compare T-PRIME with the legacy correlation-based WiFi preamble detection algorithm and ML models from prior work. T-PRIME achieves superior results on synthetically generated datasets compared to these other approaches. In the second phase, we conduct an in-depth evaluation of T-PRIME on a more realistic dataset consisting of over-the-air transmissions collected in diverse and complex RF environments. Additionally, we implement T-PRIME in the AIR-T testbed to demonstrate its accuracy and real-time inference latency.

\section{Offline Evaluation Over Synthetic Data}

\subsubsection{Dataset Description}\label{dataset description}
We used MATLAB's WLAN Waveform Generator to generate  802.11b, 802.11g, 802.11n and 802.11ax waveforms. This dataset contains baseband IQ samples according to the specifications shown in Table \ref{tab:Synthetic-Dataset}. For each protocol, we generate $2000$ Physical layer Protocol Data Units (PPDU) bursts, each comprising of multiple packets carrying a $1000$ bit random payload, saved into individual files. As 802.11b does not support a bandwidth/sampling rate of $20$MHz,  the waveform is upsampled from $11$MHz to $20$MHz to ensure  waveforms have the same channel bandwidth and ensure the problem is challenging for the classifier.

\begin{table}[t]
    \centering
    \resizebox{\columnwidth}{!}{
        \begin{tabular}{|l| c c c c c |} 
            \hline
            \textbf{Standard}     & \textbf{Bandwidth} & \textbf{MCS} & \textbf{Modulation}  & \textbf{\#PPDU/burst} & \textbf{\#IQ/burst} \\ 
            \hline
            802.11b*  & 11MHz       & QPSK     & DSSS    &  1 & 18112     \\
            802.11g  & 20MHz       & 16-QAM (½ \& ¾)     & OFDM  & 4 \& 4 & 32960     \\
            802.11n & 20MHz      & 16-QAM (½ \& ¾)  & OFDM  & 5 \& 6 & 31340       \\
            802.11ax & 20MHz      & 16-QAM (½ \& ¾)  & OFDMA  & 5 \& 6 & 31640      \\
            \hline
            
        \end{tabular}
    }
    \caption{Synthetic dataset specifications. 802.11b* is upsampled from 11MHz to 20MHz for consistency with other signals.   PPDUs per burst are reported for each coding rate.}
    \label{tab:Synthetic-Dataset}
\end{table}

\subsubsection{Training Procedure}  \label{training procedure}
Once the burst signals have been generated, we use a PyTorch data generator to train our models that operates on a per-training batch basis to dynamically load into memory each file individually and simulate transmission through a channel model at an SNR level randomly chosen. We consider the following options for (a) channel models: TGn, TGax, Rayleigh, no channel impairment; and (b)
    a random SNR level defined between $-30$ dB and $30$ dB, applied using Additive White Gaussian Noise (AWGN).
We develop a Python wrapper for Matlab's channel models standard compliant implementation and employ Model-B multi-path delay profile for TGn and TGax with device distance of $3$m, while for Rayleigh model we use a single Non-Line Of Sight (NLOS) path with an average path gain of $-3$ dB and delay of $1.5$e-$9$ seconds. %Each of the above channel and noise models can be applied at both training time, with varying randomness across epochs. Once transmissions are passed through the channel and noise simulator, they are sliced according to the model's input size and processed using its feed forward function.

Input sequences are pre-generated before training procedure for the complete set of baseband signals in the dataset. %, based on desired sequence size of each transformer model. 
Once sequence indexes are generated for each dataset burst, they are randomly shuffled and then separated into training and validation set with a split of $80$\% and $20$\%. When a mini-batch of sequences is generated at training time, the complete burst signal is retrieved and passed through a random channel model and AWGN. Finally, the input sequence is retrieved from the complete distorted signal and divided into slices as described in Fig. \ref{Fig: IQtotokens}. %This is because slices are not synchronized with the beginning of each packet and 
 Multi-path distortions thus impact the complete set of IQ samples in a batch on a per-sequence basis,  simulating realistic transmission. %Note that a new random channel instance is also generated at each channel application. 
 
 While dynamic signal augmentation is performed for both train and validation data, to evaluate the performance of our models in similar conditions we use an additional static test set, composed of $500$ bursts for each protocol following the configuration in Table \ref{tab:Synthetic-Dataset}. Such test transmissions are pre-processed by creating multiple copies of the bursts that are passed through a random channel instance for all the considered channel models, for a total of an additional $2000$ test packet bursts. Then, at test time, we create multiple sequences from each signal copy and evaluate them on discrete SNR levels ranging between $-30$ and $30$ dB in increments of $5$ dB. %We use the same training and evaluation procedure for all considered models.

%The final architecture's design, as well as the selected model and training hyperparameters, were determined through hyperparameter tuning. A grid search methodology was employed to evaluate various combinations of hyperparameters and find the optimal values. After conducting over 150 runs, we arrived at the following choices: a training duration of five epochs; a batch size of 122; and a learning rate of 2e-4.
The proposed T-PRIME design was finalized using the same training procedure above and a hyperparameters tuning phase based on a grid search on the validation set. The chosen training configuration for transformers architectures is $5$ training epochs, batch size of $122$ and learning rate $2$e$-4$. We found that a stack of $2$ transformer layers is optimal, along with two distinct sets of values for sequence length and slice length that provided nearly perfect results. The first model, named Small (SM) Transformer, utilized an input sequence of $M=24$ and tokens with a length of $S=64$. The second, Large (LG) model, employed a sequence of $M=64$ tokens with a length of $S=128$ each. The resulting number of parameters of SM and LG architectures are $1.6$M and $6.8$M, respectively.

\subsection{Comparison with state-of-the-art legacy and ML methods}

%ADD A DESCRIPTION OF COMPETITOR METHODS.

\begin{figure}
     \centering
     \begin{subfigure}[b]{0.5\textwidth}
        \centering
        \hbox{\hspace{-0.1em} \includegraphics[width=8cm]{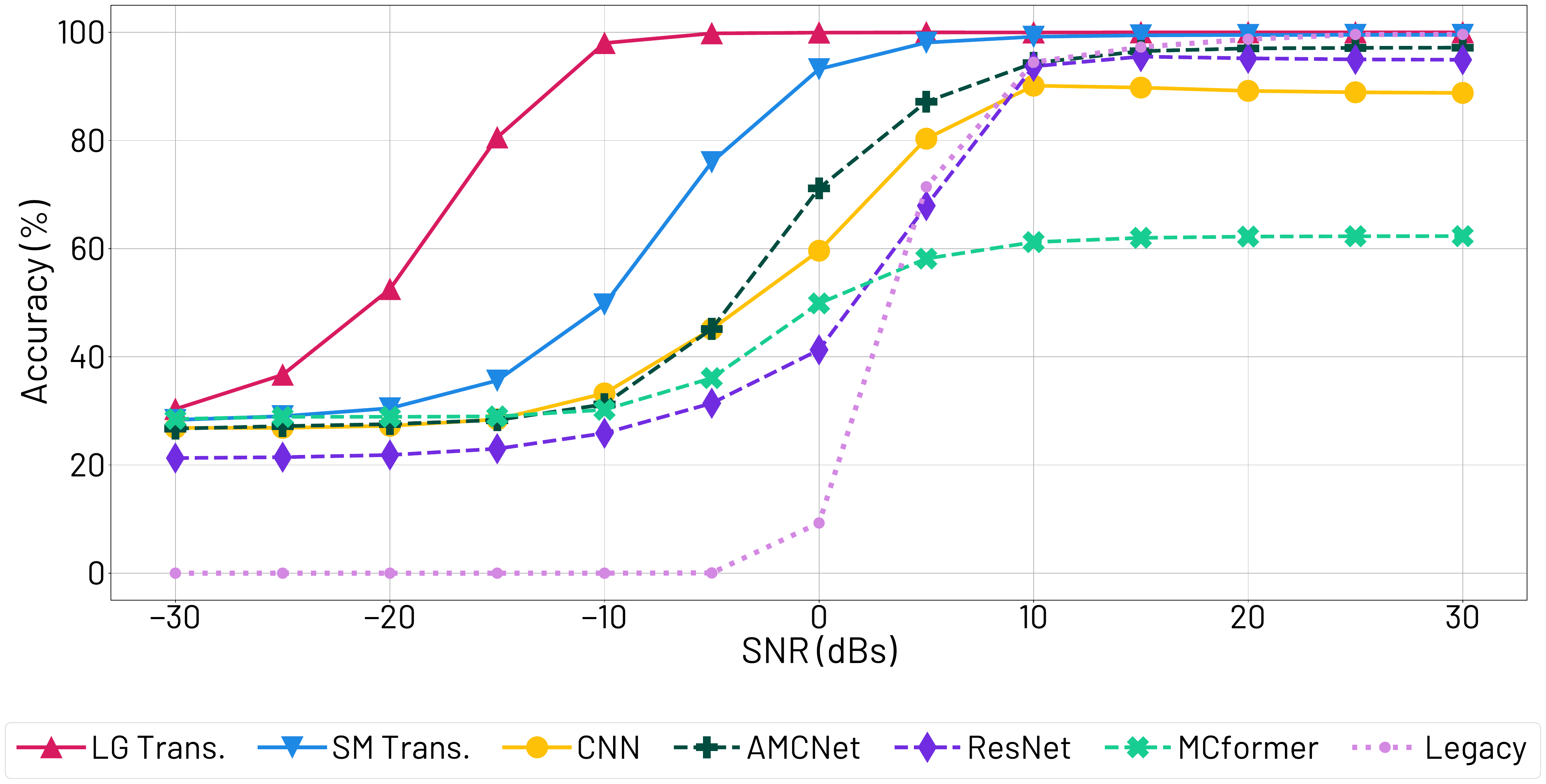}}

        \caption{}
        \label{fig:synth-model-comparison}
     \end{subfigure}
     \hfill
     \begin{subfigure}[b]{0.5\textwidth}
        \centering
        \hbox{\hspace{-0.1em} \includegraphics[width=8cm]{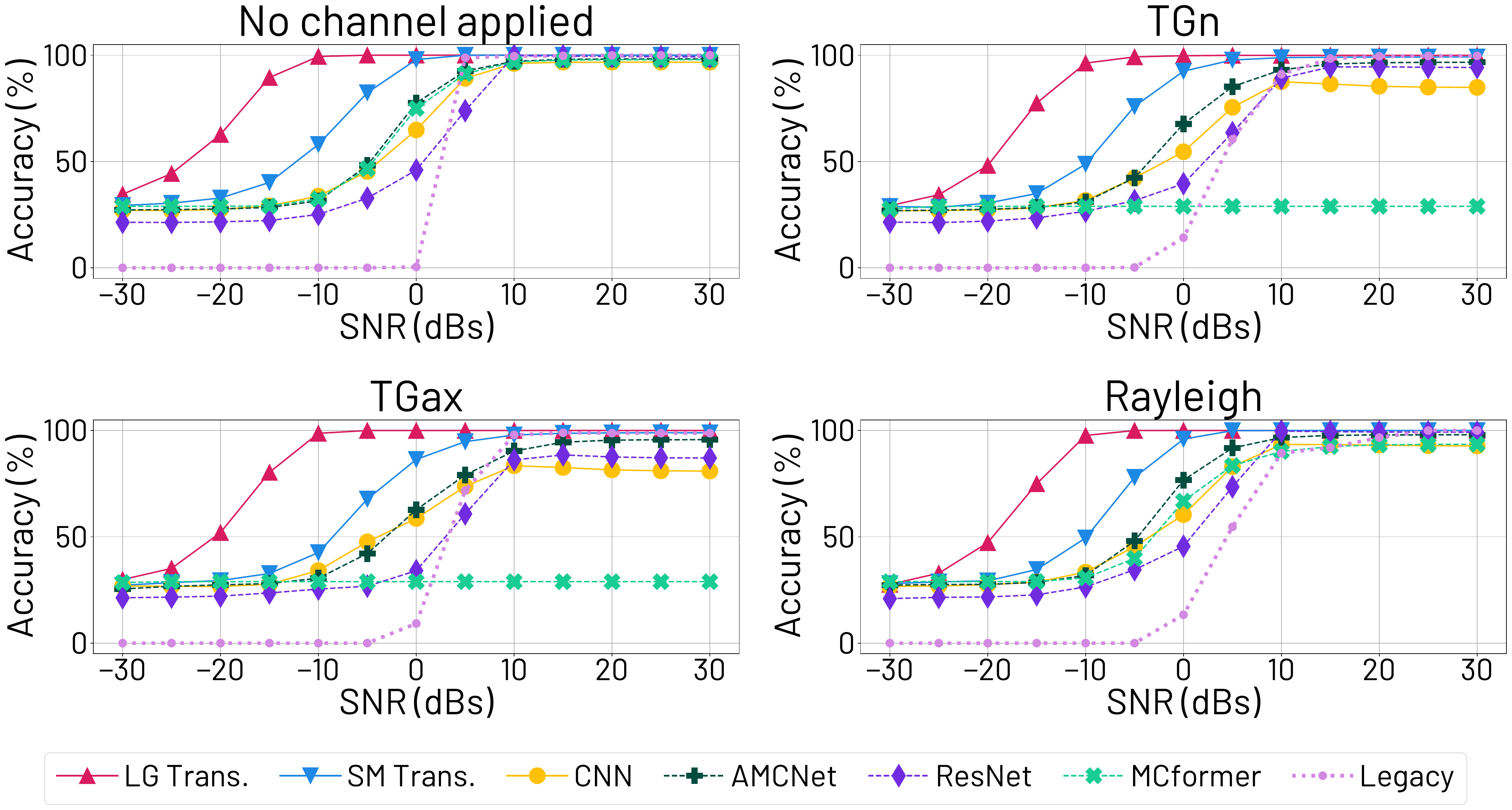}}
        \caption{}
    \label{fig:synth-model-comparison-bkdown}
     \end{subfigure}
     
        \caption{(a) Comparison between SM, LG Transformer-based architectures and state-of-the-art signal classification models tested on all channel models and different SNR conditions. The LG Transformer achieves the best overall accuracy and it retains high accuracy (i.e. $> 98\%$) for SNR as low as -10 dB. All models improve under optimal hyperparameter search (see Appendix), but the LG Transformer still outperforms all other methods. (b) Performance comparison for each individual channel model tested during simulation.}
        \label{fig:tprime-competitors-caccuracy}
\end{figure}

%\subsubsection{Comparison with Legacy Method}
\label{sec:legacy_performance_simulation}
We first show how T-PRIME approach can effectively improve the state-of-the-art over legacy approaches. Specifically, we use the same static test set described in Sec. \ref{dataset description} for a simulation study that processes each test transmission using the correlation-based procedure specified in the standard \cite{9363693}. We employ a set of multiple decoders provided by Matlab that use the legacy method to detect the following preambles formats: 
1) VHT (i.e. 802.11n), 2) HE-SU (i.e. 802.11ax), 3) Non-HT (i.e. 802.11g/b).
We then process our test transmissions through these functions for each channel model considered. %, using the same simulation parameters used while training on synthetic data. 
For each test signal, we simulate $100$ transmissions over random channel instances for each channel model and SNR levels considered in our evaluation procedure described before. To report the accuracy of legacy procedures, %we compute the inverse Packet-Error-Rate (i.e. 1 - PER) 
we consider the inverse Packet-Detection-Rate (i.e. $1$ - PDR), defined as the ratio of correct preamble format detections over the total number of simulated packets transmissions, and compare it with accuracy of proposed architectures under the same statistical channel and SNR conditions. We evince from Fig. \ref{fig:synth-model-comparison} that both our T-PRIME models drastically outperform the legacy detection mechanism, achieving near-perfect detection accuracy already at SNR $= 0$ dB, while traditional methods completely fail to operate under such circumstances. This demonstrates how these models are still able to extract and learn powerful time-domain dependencies from received signals even in presence of strong noise or low reception power.
After establishing superiority of our approach over legacy methods, we compare our model to other state-of-the-art ML methods for wireless signal classification. While our problem formulation differs from modulation classification, it serves as a relevant example of architectures processing sequences of raw IQ samples for similar tasks. We compare T-PRIME against three state-of-the-art models in this domain: ResNet \cite{8267032}, AMCNet \cite{zhang2023amc}, and MCformer \cite{hamidi2021mcformer}, using the hyperparameters reported as optimal by their respective authors; we also present supplementary findings obtained through additional hyperparameter exploration in Appendix.
% {\color{blue} We also conducted supplementary experiments, including an extensive hyperparameter tuning campaign, to further compare the performance of our proposed solutions with state-of-the-art signal classification models. While other machine learning models exhibited improved performance with optimized hyperparameters, our results demonstrate that T-PRIME consistently outperforms all competing solutions, even after hyperparameter tuning. Additional results and further details can be found in the Appendix.}
% {\color{cyan} We also conducted hyperparameter tuning experiments to further compare the performance of our proposed solutions with state-of-the-art signal classification models. While other machine learning models exhibited improved performance with optimized hyperparameters, our results demonstrate that T-PRIME still consistently outperforms all competing solutions, even after hyperparameter tuning. Additional results and further details can be found in the Appendix.}
For further comparison, we trained a 1D CNN model inspired by \cite{o2016convolutional} with a wider input size and slightly larger set of parameters. We used a common training configuration for all models with a batch size of 512, $M=1$ for input generation, \textit{CrossEntropyLoss} as the loss function, and a learning rate of $0.001$ with exponential decay to $0.0001$ when a plateau in the validation loss is reached. Table \ref{tab:model-size-comparison} lists all the models we compared, along with their input size and total number of parameters.

\begin{table}[]
    \centering

    % \begin{tabular}{|l|ll|}
    % \hline
    % \textbf{Model} &\textbf{ \# Param.} & \textbf{Input size} ($N \times S$) \\ \hline
    % T-PRIME LG & 6.8M & $64 \times 128$ \\
    % T-PRIME SM & 1.6M & $24 \times 64$ \\
    % 1D CNN \cite{o2016convolutional} & 4.1M & $1 \times 512$ \\
    % ResNet \cite{8267032} & 162K & $1 \times 1024$ \\
    % AMCNet \cite{zhang2023amc} & 462K & $1 \times 128$ \\
    % MCformer \cite{hamidi2021mcformer} & 78K & $1 \times 128$ \\ \hline
    % \end{tabular}
    \begin{tabular}{|l|ll|}
    \hline
    \textbf{Model} & \textbf{Input size} ($M \times S$) & \textbf{\# Parameters} \\ \hline
    T-PRIME LG & $64 \times 128$ & 6.8M \\
    T-PRIME SM & $24 \times 64$ & 1.6M \\
    1D CNN \cite{o2016convolutional} & $1 \times 512$ & 4.1M \\
    ResNet \cite{8267032} & $1 \times 1024$ & 162K \\
    AMCNet \cite{zhang2023amc} & $1 \times 128$ & 462K \\
    MCformer \cite{hamidi2021mcformer} & $1 \times 128$ & 78K \\ \hline
\end{tabular}
    \caption{Comparison between proposed architectures and state-of-the-art MCS classification architectures.}
    \label{tab:model-size-comparison}
\end{table}

%Once we established the superiority of our approach over legacy methods, we compare our model to other state-of-the-art machine learning methods for wireless signal classification.
%As noted in Sec. \ref{background-ml}, there is a growing body of work using ML models to perform modulation classification. While our problem formulation highlights how MCS classification alone is not enough to correctly classify transmission protocols, these models constitute the closest example of architectures used to process sequences of raw IQ samples for similar classification task. Therefore, we decide to compare the performance of T-PRIME against three state of the art models in this domain: ResNet \cite{8267032}, AMCNet \cite{zhang2023amc}, and MCformer \cite{hamidi2021mcformer} using optimal hyperparameters as reported by these works. For further comparison, we also trained a 1D CNN model inspired by \cite{o2016convolutional} with a wider input size and  slightly larger set of parameters. %We used the same training procedure described in Sec. \ref{training procedure} for all competitor models. 
%We use a common training configuration for all models using batch size of 512, $N = 1$ for input generation, \textit{CrossEntropyLoss} as loss function and learning rate of 0.001 with exponential decay to 0.0001 when a plateau in the validation loss is encountered. Table \ref{tab:model-size-comparison} lists all models we compared along with the input size and total number of parameters.

Fig. \ref{fig:tprime-competitors-caccuracy} illustrates that both T-PRIME LG and T-PRIME SM consistently outperform all other methods across the entire range of evaluated signal-to-noise ratios (SNRs). Notably, T-PRIME LG achieves a remarkable $>60$\% improvement over all other models at SNR $= -10$ dB, achieving 98\% accuracy. %Despite having fewer parameters, T-PRIME SM still outperforms all other models. 
From these experiments, two key points can be deduced: 1) T-PRIME's performance benefits from a larger parameter space, following the trend of Large-Language-Models (LLMs) that use substantial parameter capacity to learn complex relationships \cite{wei2022emergent}; 2) Our problem significantly differs from MCS classification, as the results suggest that larger amounts of temporal sequences are required to accurately identify the complex and less localized patterns in real-world wireless protocols, especially when protocols share the same MCS configurations. Finally, hyperparameter search improves all models (see Appendix); while, in this case, AMCNet surpasses the SM Transformer for low SNR, nevertheless the LG Transformer still consistently outperforms all competitors.

\section{Real-Time Implementation on  AIR-T SDR}

\begin{figure}[tb]
    \centering
    \includegraphics[width=.75\linewidth]{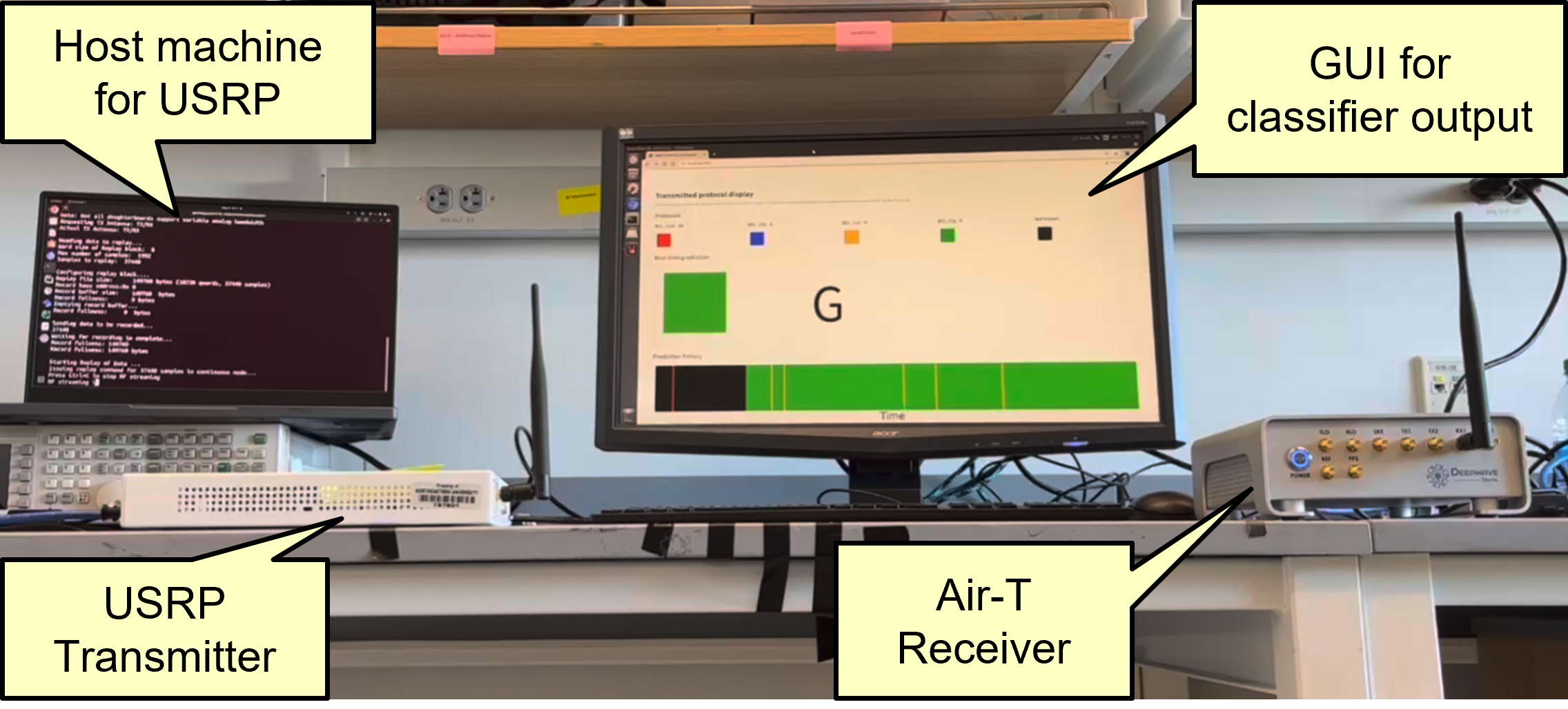}
    \caption{T-PRIME running real-time classification in an AIR-T receiver. 802.11g protocol is shown to be correctly detected in this instance.}
    \label{Fig: system pic}
\end{figure}

SDR-based systems offer flexibility in the receiver chain and seamless integration with AI/ML modules. However, they can introduce higher latency due to challenges like I/O and buffer management, software dependencies, and distribution of computing resources for signal processing tasks. To explore these systems' challenges and showcase T-PRIME's real-time efficiency, we implemented it on Deepwave Digital's AIR-T model AIR7101, as shown in Fig \ref{Fig: system pic}. The AIR-T, built on NVIDIA's Jetson TX2 module, includes a CPU with $4$ ARM A-57 cores, $2$ ARM Denver2 cores, and a Pascal $256$-core GPU with $8$ GB of shared memory \cite{AIR-T}. This design resolves traditional SDR system issues by integrating resources on one platform, including SDR with FPGA, networking, periphery connections, an Ubuntu-based OS, and open source APIs.

%SDR based systems naturally support more flexible receiver chain and integrate well with AI/ML modules. However, they generally incur a cost of higher latency due to I/O and buffer management challenges, software dependency complexities, and distribution of computing resources for the signal processing tasks. In order to analyze the systems level challenges of performing machine learning for protocol detection on edge devices, and demonstrate the feasibility and real-time efficiency of T-PRIME, we implemented it on Deepwave Digital's AIR-T model AIR7101.  Fig \ref{Fig: system pic} shows T-PRIME making real time predictions in a lab environment.
%To support our goal of real time protocol detection using network edge devices, we implemented AIR-TPC on Deepwave Digital's AIR-T model AIR7101. 

\begin{figure}[tb]
    \centering
    \includegraphics[width=.95\linewidth]{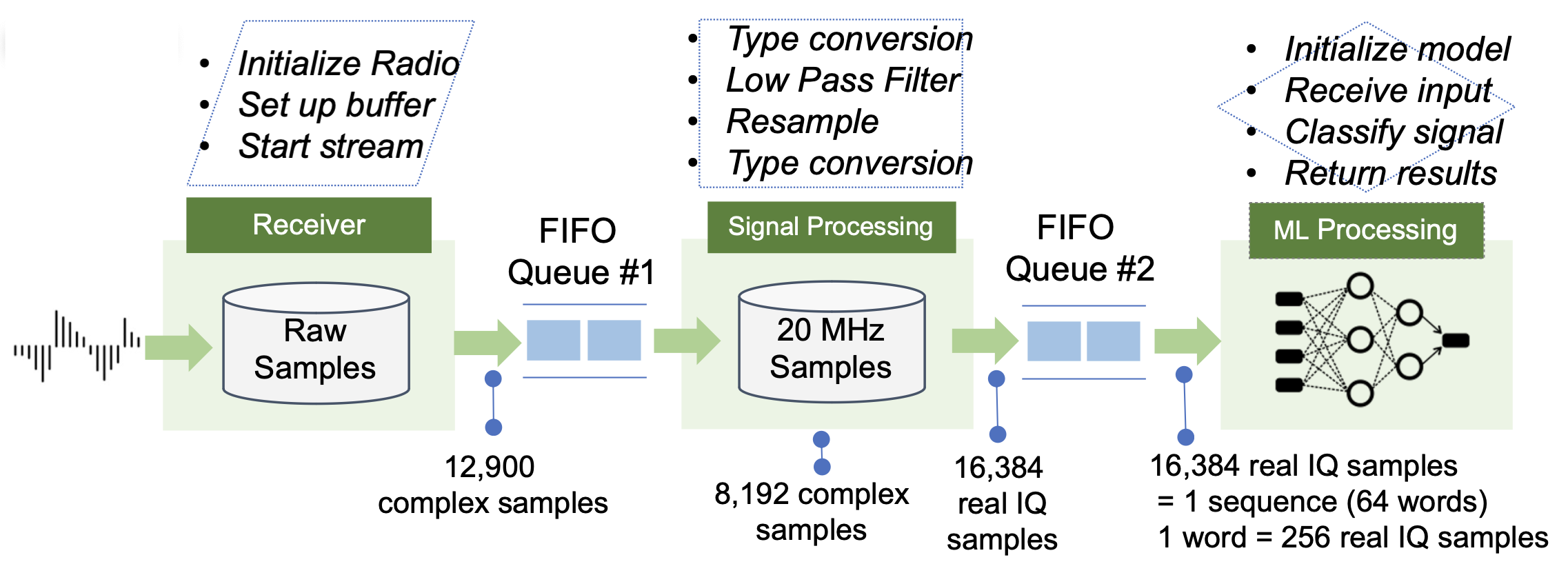}
    \caption{T-PRIME pipeline overview showing three steps of the receiver, signal processing and ML processing modular blocks.}
    \label{Fig: pipeline_overview}
\end{figure}

%\subsection{AIR-T and Testbed Implementation}
For real-time protocol classification, low latency is crucial. Hence, we designed a pipeline (Fig. \ref{Fig: pipeline_overview}) with three primary functions running in parallel: receiver, signal processing, and ML inference. This parallelization removes concurrent dependencies and reduces prediction delay. It also allows individual function modification or improvement without affecting others. For instance, we deployed the same Transformer model using two frameworks (PyTorch and TensorRT) transparently. Each thread includes timing functions to calculate the average time for function completion. Additionally, we established two thread-safe FIFO queues, \texttt{q1} between receiver and signal processing threads, and \texttt{q2} between signal processing and ML inference threads. Below, we briefly describe each of these three threads;  see also our code~\cite{tprime_github} for further details.

\subsubsection{Receiver Thread} The main function of the receiver thread is to capture the raw OTA signals. The receiver thread defines several parameters such as sample rate ($31.25$ MHz), gain mode, and frequency for the AIR-T receiver using SoapySDR APIs. After initializing the SDR, it creates a data buffer to receive samples from the SDR. After these one time initialization steps are complete, the receive thread enters a continuous loop that reads the buffer, ensures it received the expected number of samples, and writes the vector of samples to \texttt{q1} if there is space available. 

\subsubsection{Signal Processing Thread} The signal processing thread's main function is to down-sample the received signal to $20$ MHz and convert it to the format required by the ML inference thread. After reading the raw sample vector from \texttt{q1}, the thread converts the data type from interleaved shorts to \texttt{numpy.complex64}, normalized to the range of $[-1, 1]$. Then, %a low pass filter and re-sampling function 
we down-sample the signal from $31.25$ MHz to $20$ MHz, the standard channel bandwidth used in 802.11 WiFi protocols. Finally, the complex vector is converted into a real vector by interleaving the I and Q (real and imaginary) components, as shown in Fig. \ref{Fig: IQtotokens}. If space is available, the thread writes this vector to \texttt{q2}.

%The primary objective of this thread is to down-sample the received signal to 20 MHz and convert the data format to what is expected by the ML inference thread. After the signal processing thread reads the raw sample vector from \texttt{q1}, it converts the received sample data type from interleaved shorts to numpy.complex64 normalized to the range $[-1, 1]$. Next, we use a low pass filter and re-sampling function to down-sample the signal from 31.25 MHz to 20 MHz, the standard channel bandwidth used in 802.11 WiFi protocols. Finally, we convert this complex vector into a real vector by interleaving the I and Q (real and imaginary) components. This final step is visualized as the flattening of IQ samples in Fig. \ref{Fig: IQtotokens}. If there is space available the thread writes this vector to \texttt{q2}.

\subsubsection{ML Inference Thread} This thread is the core of our pipeline, performing the actual ML inference on the input samples. We deployed two different versions of this thread; a PyTorch version and a TensorRT version. TensorRT \cite{nvidiaQuickStart} uses techniques such as quantization, layer and tensor fusion, kernel tuning to optimize DL inference. Regardless of the optimization used, the thread follows the same process. After loading the model, it enters a continuous loop. If there is data available in \texttt{q2} it reads that into a tensor to use as the model input,  classifies, and appends the classification label to a text file.

%\noindent\textbf{Signal Power Normalization.}
To handle varying signal power caused by propagation effects, we incorporate a power normalization block in the ML inference thread. This function normalizes the total power of received signals to a nominal power of 1W.
%To deal with varying signal power introduced by propagation effects, we  add a power normalization block to the ML inference thread, before the received IQ samples are passed as input to the neural network. This function normalizes the total power of the received signals to a nominal power of $1$W. 
Given a generic discrete complex signal $\mathbf{s} \in \mathbb{C}^N$ of $N$ samples, we can obtain its normalized version $\mathbf{\hat{s}}$ as follows:
%\begin{equation}
$\mathbf{\hat{s}}={\mathbf{s}}/{\sqrt{\frac{\sum_{i=1}^N\left|s_i\right|}{N}}}$
%\end{equation}
where the denominator is the Root-Mean-Square (RMS) value of the incoming signal. This block makes our model more robust to power and SNR variations, eliminating the need to collect data samples and train our models under multiple power profiles. %However, it also amplifies noisy patterns when only background noise is observed.

\section{T-PRIME AIR-T Testbed  Evaluation}

We evaluated the performance of our T-Prime implementation through: \textit{(a)} the creation of an Over-the-Air (OTA) dataset, collected from the deployment of the testbed in five different locations, and use this dataset to conduct additional offline evaluation of T-Prime, \textit{(b)} power and latency profile characterization of our implementation, \textit{(c)} real-time, live tests of a deployment of our entire pipeline evaluating our models in a new environment. We describe each of these experiments below.

\subsection{Experimental Setup}
\subsubsection{Over-the-Air (OTA) Experiments}
\label{OTA_dataset_description} In order to evaluate the performance of our proposed approach in a real-world setting, we collect OTA transmissions in a variety of conditions, including different indoor scenarios and transmission power levels. To collect this data, we re-use the same receiver and signal processing blocks. Instead of passing the resulting vectors to the ML inference block, we write them to a file for offline training and testing.

\noindent $\bullet$ \textbf{Dataset collection.} 
We collected a large OTA dataset, providing a robust collection of real-world wireless channel conditions. Table \ref{tab:OTAdata} summarizes the environments for each collection set. Room A is a large open classroom, while rooms B and C are small lab rooms. The level of interference was assessed based on visible WiFi access points and other known transmitters in the $2.4$ GHz band. The level of scatter considers the ratio of room dimensions to the space used by equipment and furniture. Each room's data was collected on two separate days (\_1 vs \_2). We also collected noise/background samples to measure the noise floor and test classifier performance when no active transmission occurred.

\begin{table}[!t]
    \centering
    \resizebox{\columnwidth}{!}{
    \begin{tabular}{|llll|rrrrr|}
        \hline
        \multicolumn{1}{|l|}{\textbf{Dataset}} & \multicolumn{1}{c|}{\textbf{Distance}} & \multicolumn{1}{c|}{\textbf{Scatter}} & \multicolumn{1}{c|}{\textbf{Interference}} & \multicolumn{5}{c|}{\textbf{\#Captures per protocol}} \\ \hline
        \multicolumn{1}{|l|}{} &  &  &  & \multicolumn{1}{l}{ax} & \multicolumn{1}{l}{b} & \multicolumn{1}{l}{n} & \multicolumn{1}{l}{g} & \multicolumn{1}{l|}{noise} \\ \hline
        \multicolumn{1}{|l|}{RM\_A\_1} & 3m & Low & Low & 200 & 206 & 200 & 200 & 0 \\
        \multicolumn{1}{|l|}{RM\_A\_2} & 2m & Low & Medium & 200 & 200 & 200 & 200 & 0 \\
        \multicolumn{1}{|l|}{RM\_B\_1} & 1m & Med & Medium & 73 & 200 & 200 & 200 & 0 \\
        \multicolumn{1}{|l|}{RM\_B\_2} & 1m & Med & Medium & 500 & 500 & 500 & 500 & 500 \\
        \multicolumn{1}{|l|}{RM\_C\_1} & 1m & High & High & 100 & 100 & 100 & 100 & 0 \\
        \multicolumn{1}{|l|}{RM\_C\_2} & 1m & High & High & 500 & 500 & 500 & 500 & 100 \\ \hline
\end{tabular}
    }
    \caption{OTA single-protocol datasets comparison. Each capture file contains 198080 IQ samples - approximately 10 \unit{\milli\second} duration. %{\color{red} Maybe use room pictures from the slides? and include this table  as a "subfigure".}
    }
    \label{tab:OTAdata}
\end{table}
%For each protocol, we generate a small set of standard compliant WiFi frames, composed of both preamble and random data payload in baseband. Once this signal is configured, we continuously transmit the given signal and capture it at the receiver side after it travels through the wireless medium. For all our data collections, we employ Ettus USRP X310 to transmit our baseband signals using $30$ dBm trasmission power (unless specified otherwise) and collect it with the modified T-PRIME. Both transmitter and receiver are configured to set their central frequency as $f_c = 2.442$ GHz, i.e. WiFi Channel 7. 
%Unless otherwise noted, both our training and testing phases include a power normalization pre-processing step.

For each protocol, we generated standard compliant WiFi frames, including preamble and random data payload in baseband. We continuously transmitted these signals and captured them at the receiver side after traveling through the wireless medium. Each capture is approximately $10$ \unit{\milli\second} in duration. The transmitter was an Ettus USRP X310 with $30$ dBm transmission power, and both transmitter and receiver were set to a central frequency of $f_c = 2.442$ GHz, corresponding to WiFi Channel 7. In both training and testing phases, we applied  power normalization, unless otherwise specified.

\noindent $\bullet$ \textbf{Single-protocol  OTA dataset.} The above data collection results in an offline OTA dataset summarized in Table \ref{tab:OTAdata}. We split this into a training and test datasets in two ways. In our \emph{time-split}, for each day of data collection, we consider the first $80$\% of  samples collected per protocol as a training set, and the remaining $20$\% as a test set. In our \emph{scenario-split},  we perform a cross-validation study across all the location-specific datasets. We train our classifier on all OTA collections except one, which we use in its entirety to test the resulting model.  In both cases the test data  channel conditions are unseen during the training phase. 

\noindent $\bullet$ \textbf{Overlapping-protocol  OTA datasets.} We stress-tested T-PRIME in the presence of spectral overlaps, a scenario where traditional preamble detection techniques struggle due to frame collisions corrupting the preamble structure. In RM\_C, we collected two additional datasets with signals transmitted concurrently, overlapping at $25\%$ and $50\%$ spectral ratios. We considered six different overlapping protocol configurations (see Table \ref{tab:wifi_combinations}). The datasets differed in the observed wireless spectrum resolution, achieved by varying the receiver sampling rate and central frequency.

\begin{table}[!t]
    \centering
    \begin{tabular}{|c|c|c|}
        \hline
        \textbf{Configuration} & \textbf{Incumbent} & \textbf{Interferer} \\
        \hline
        C1 & 802.11g & 802.11n \\
        C2 & 802.11b & 802.11ax \\
        C3 & 802.11b & 802.11g \\
        C4 & 802.11b & 802.11n \\
        C5 & 802.11g & 802.11ax \\
        C6 & 802.11n & 802.11ax \\
        \hline
    \end{tabular}
    \caption{Protocols configuration for overlapping cases.}
    \label{tab:wifi_combinations}
\end{table}

\begin{figure}
     \centering
     \begin{subfigure}[b]{0.22\textwidth}
        \centering
        \hbox{\hspace{0em} \includegraphics[width=4.1cm]{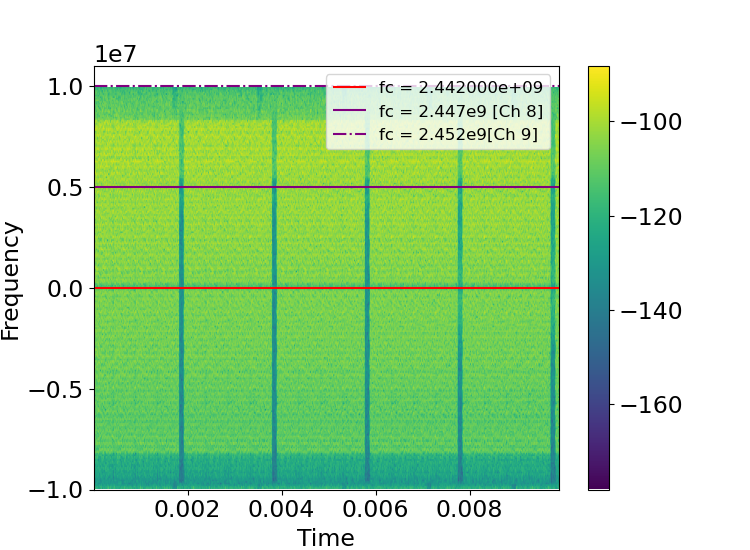}}

        \caption{$f_s^{\text{Rx}} = 20$, $R = 25\%$}
        \label{fig:overlap25_gax_2}
     \end{subfigure}
     ~
     \begin{subfigure}[b]{0.22\textwidth}
        \centering
        \hbox{\hspace{0em} \includegraphics[width=4.1cm]{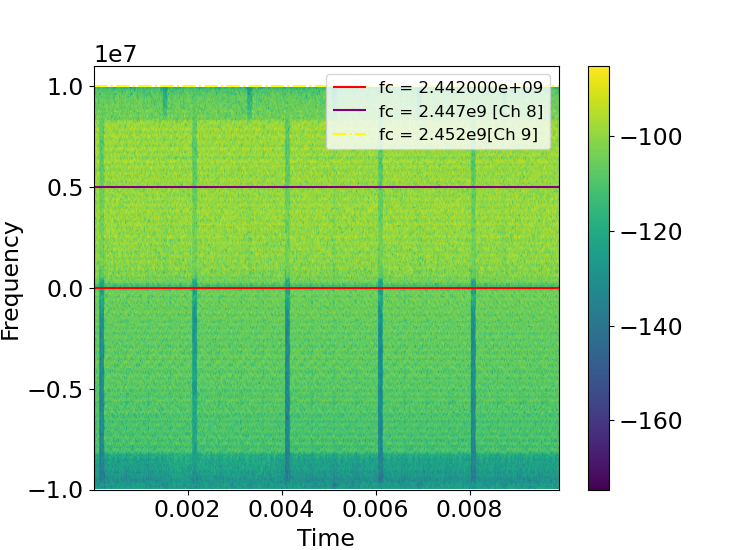}}
        \caption{$f_s^{\text{Rx}} = 20$, $R = 50\%$}
    \label{fig:overlap50_gax_2}
     \end{subfigure}
     \begin{subfigure}[b]{0.22\textwidth}
        \centering
        \hbox{\hspace{0em} \includegraphics[width=4.1cm]{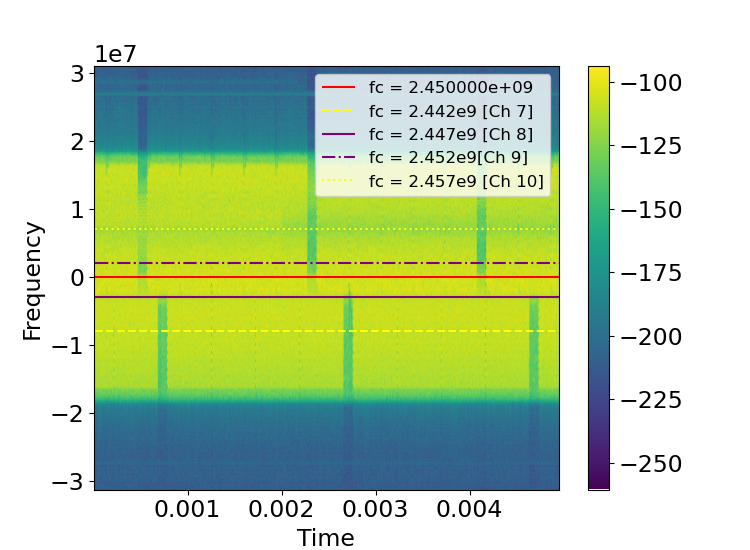}}

        \caption{$f_s^{\text{Rx}} = 62.5$, $R = 25\%$}
        \label{fig:overlap25_gax_1}
     \end{subfigure}
     ~
     \begin{subfigure}[b]{0.22\textwidth}
        \centering
        \hbox{\hspace{0em} \includegraphics[width=4.1cm]{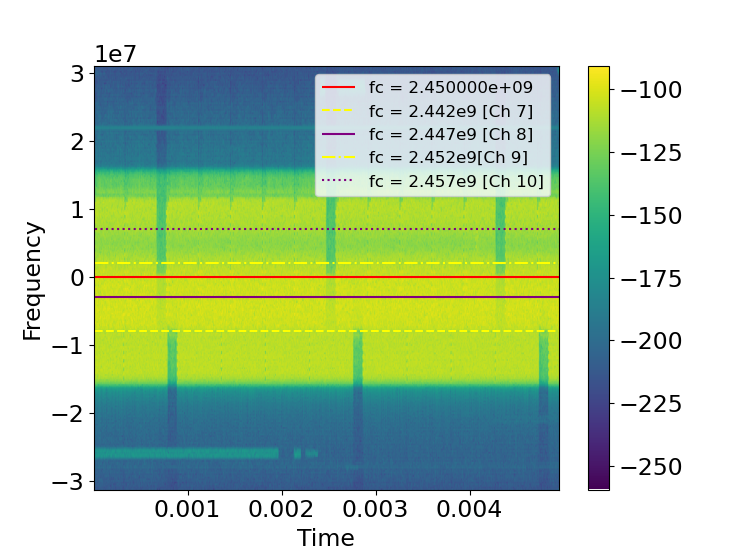}}
        \caption{$f_s^{\text{Rx}} = 62.5$, $R = 50\%$}
    \label{fig:overlap50_gax_1}
     \end{subfigure}
     
        \caption{ Example spectrograms of recorded transmissions used in the interference dataset. Samples are shown for each configurations of overlapping rate $R$ and sampling rate $f_s^{\text{Rx}}$ expressed in MHz.}
        \label{fig:overlap_samples}
\end{figure}

Our goal was to test if our Transformer classifier could recognize both incumbent signals and the protocol of the interfering transmission using raw IQ time-domain samples. We defined two additional datasets with different receiver configurations: O1) the incumbent and receiver tuned to the same WiFi channel (WiFi Ch. 7) and sharing the same bandwidth, and O2) the receiver observing a wider bandwidth and tuned to an arbitrary frequency ($2.45$ GHz) to observe both overlapping signals in the frequency domain.
Table \ref{tab:OTA_overlap_dsinfo} summarizes the parameters used for the overlapping data collections, and Fig. \ref{fig:overlap_samples} visualizes each receiver configuration with C5 type (802.11g / 802.11ax) transmissions as an example. 

\begin{table}[!t]

    \centering
    \resizebox{\columnwidth}{!}{
        \begin{tabular}{|c|c|cccccc|} % Note the additional two 'c's in the argument of tabular
            \hline
            \textbf{Dataset} & $R$ & \textbf{$f_s^{\text{Rx}}$} & \textbf{$f_c^{\text{Rx}}$} & \textbf{$f_c^{\text{Tx1}}$}& \textbf{$f_c^{\text{Tx2}}$} & \textbf{\#Capt./Config} & \textbf{\#IQ/Capt.} \\ % Added the two new headers
            \hline
           RM\_C\_O1 & 25\% & 20 & 2.442 & 2.442 & 2.457 & 200 & 198080 \\
            RM\_C\_O1 & 50\% & 20 & 2.442 & 2.442 & 2.452 & 200 & 198080\\
            \hline
            RM\_C\_O2 & 25\% & 62.5 & 2.45 & 2.442 & 2.457 & 200 & 309500 \\
            RM\_C\_O2 & 50\% & 62.5 & 2.45 & 2.442 & 2.452 & 200 & 309500 \\
            \hline
        \end{tabular}
    }
    \caption{ OTA overlapping dataset  parameters. $f_s^{\text{Rx}}$ (expressed in MHz) is receiver sampling rate. $f_c^{\text{Rx}}$, $f_c^{\text{Tx1}}$ and $f_c^{\text{Tx2}}$ (expressed in GHz) are receiver central frequency, central frequencies of first transmitter (incumbent) and central frequency of second transmitter, respectively.}
    \label{tab:OTA_overlap_dsinfo}
    
\end{table}

We adapted our classifier from a single to multi-label output by replacing the last \textit{LogSoftMax} activation function with  \textit{Sigmoid} at each output neuron and using \textit{BCELoss} as loss function. We trained our classifier on all collected OTA transmissions, including single protocol and overlapping ones, using the \emph{time-split} method with the same $80$/$20$ split for training/testing. %We collect over 200 file captures for each protocol and overlap ratio configuration, each file consisting of approximately 10 \unit{\milli\second} worth of complex IQ samples. 
Note that now we include  both non-overlapping and overlapping cases, and signals with different sampling rates (e.g. \_O2 datasets) in the training set. This challenging dataset forces the Transformer to learn different patterns and ultimately classify the protocols irrespective of their sampling rate.

\subsubsection{Latency and Power Profile Experiments}
The AIR-T has $6$ power profiles, providing several different combinations of CPUs enabled and various clock rates for the CPUs and GPU.
We performed a thorough analysis of all $6$  configurations to understand the constraints of our system along with the benefit and cost of each option. %For each configuration, 
We used NVIDIA's JTOP utility to monitor power consumed while T-PRIME performed real-time protocol classification. 

\subsubsection{Real-Time Experiments}
We also evaluate the accuracy of T-PRIME LG in classifying OTA signals in real time. For this experiment we train T-PRIME using the entire single protocol OTA dataset (Table \ref{tab:OTAdata}). We deploy T-PRIME in two new locations, never seen during training. In both rooms, the transmitter is approximately $3$m from the T-PRIME receiver. We sequentially generate and transmit each of the four 802.11 protocols in the same manner as described in Sec. \ref{OTA_dataset_description}. While continuously transmitting a given protocol, we capture $1001$ live predictions made by T-PRIME. Each experiment lasts approximately $10$s.

\subsection{Results}

\subsubsection{OTA Dataset Evaluation}
First, we show the performance of our classifier on OTA signals.

%While this block amplifies noisy patterns when only background noise is observed, leading to the possibility of false detections, it makes our model more robust to power and SNR variations, eliminating the need to collect data samples and train our models under multiple power profiles. 

\begin{figure}[!t]
     \centering
     \begin{subfigure}[b]{0.22\textwidth}
        \centering
        \hbox{\hspace{0em} \includegraphics[width=4cm]{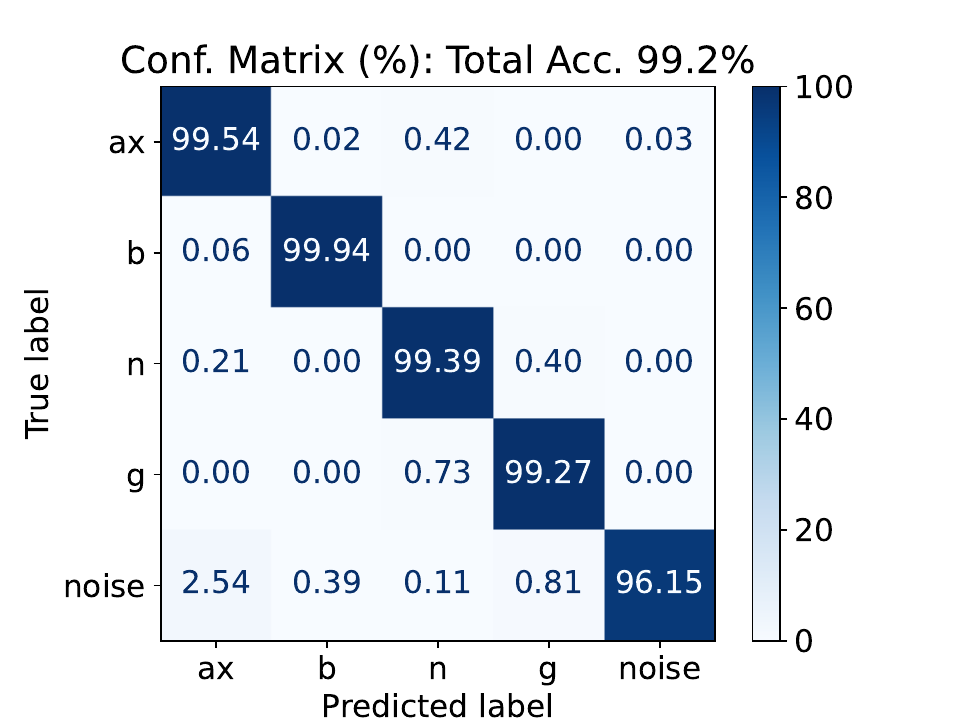}}

        \caption{LG model}
        \label{fig:OTA-LG-avg}
     \end{subfigure}
     ~
     \begin{subfigure}[b]{0.22\textwidth}
        \centering
        \hbox{\hspace{0em} \includegraphics[width=4cm]{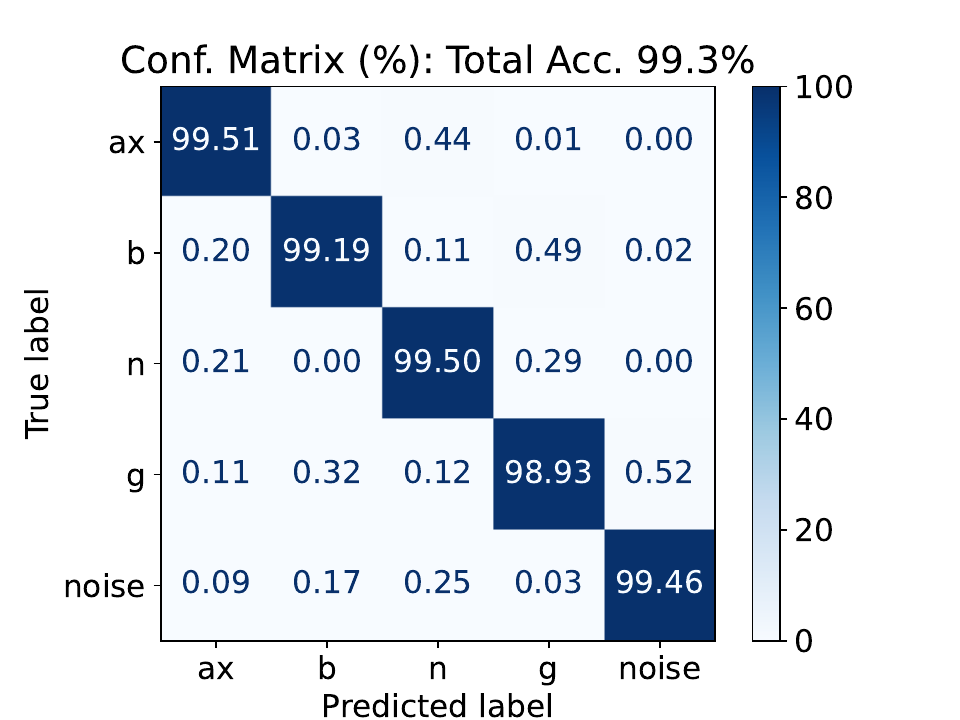}}
        \caption{SM model}
    \label{fig:OTA-LG-avg}
     \end{subfigure}
     
        \caption{ WiFi protocol classification performance on the complete OTA test dataset for single transmissions.  Comparison between Large (LG) and Small (SM) Transformer models. }
        \label{fig:LG-SM-avg-OTA}
\end{figure}

\begin{figure}
     \centering
     \begin{subfigure}[b]{0.15\textwidth}
        \centering
        \hbox{\hspace{0em} \includegraphics[width=3.5cm]{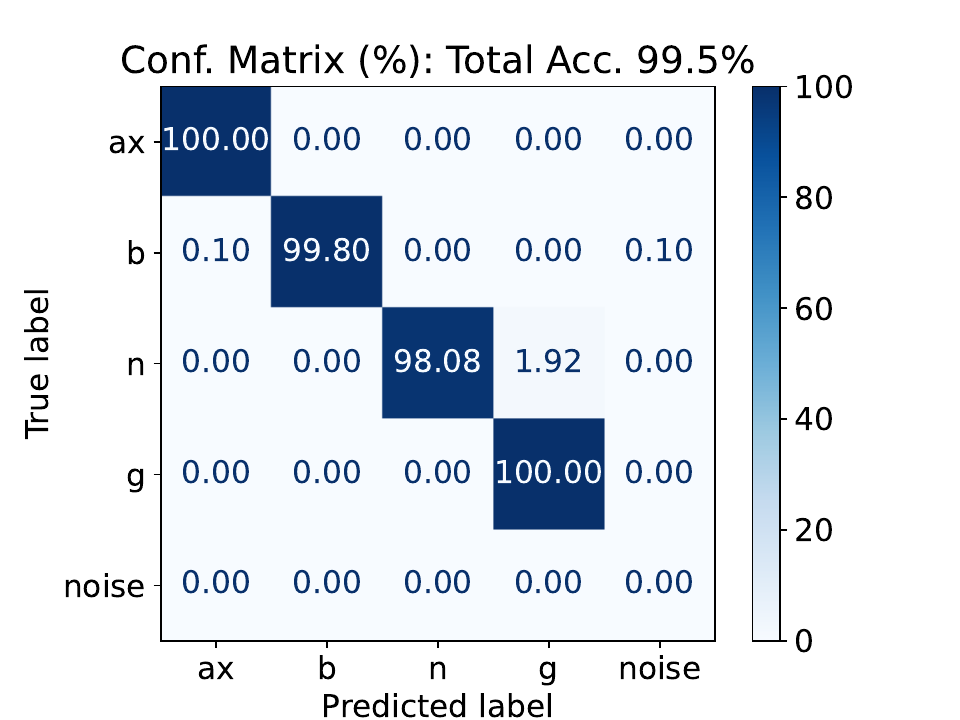}}
        \caption{RM\_A\_1}
        \label{fig:cv-RM_142_1}
     \end{subfigure}
     ~
      \begin{subfigure}[b]{0.15\textwidth}
        \centering
        \hbox{\hspace{0em} \includegraphics[width=3.5cm]{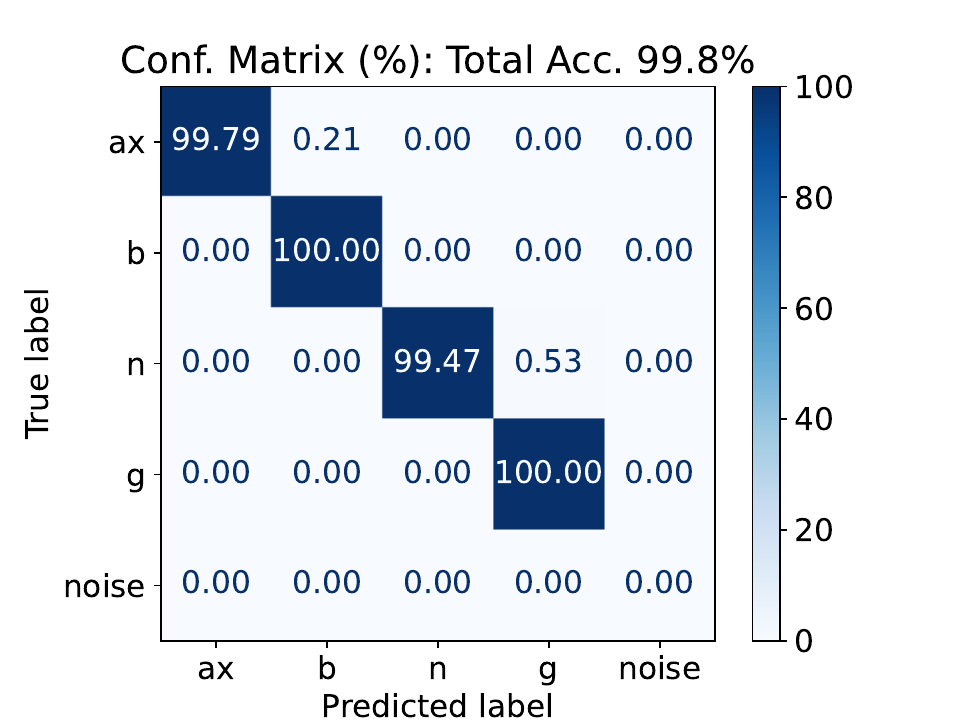}}
        \caption{RM\_A\_2}
        \label{fig:cv-RM_142_2}
     \end{subfigure}
     ~
     \begin{subfigure}[b]{0.15\textwidth}
        \centering
        \hbox{\hspace{0em} \includegraphics[width=3.5cm]{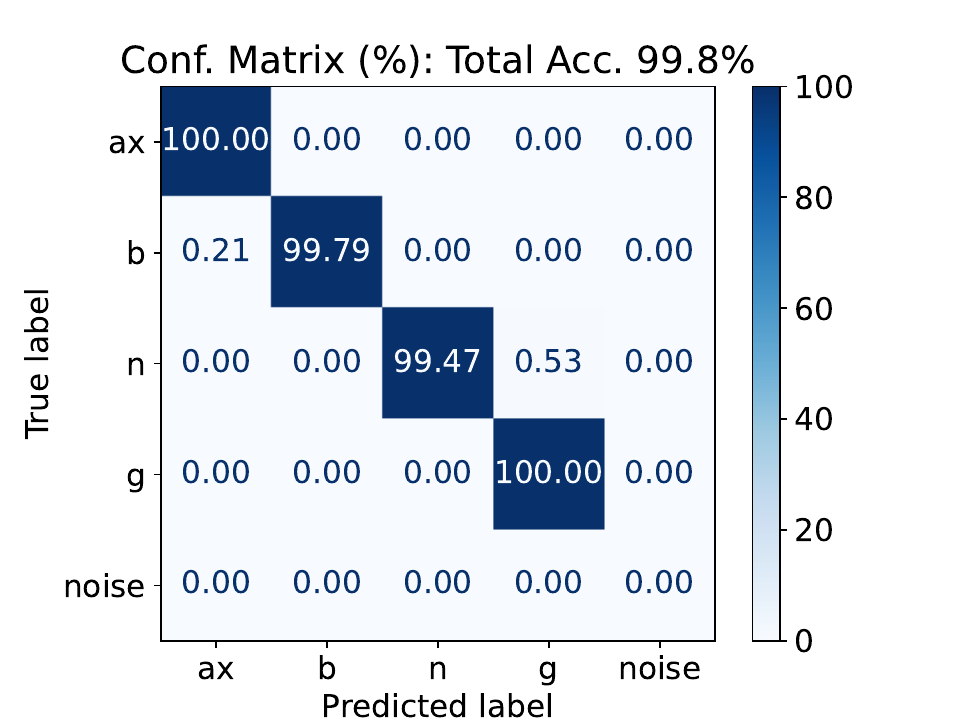}}
        \caption{RM\_B\_1}
        \label{fig:cv-RM_572C_1}
     \end{subfigure}
     ~
     \begin{subfigure}[b]{0.15\textwidth}
        \centering
        \hbox{\hspace{0em} \includegraphics[width=3.5cm]{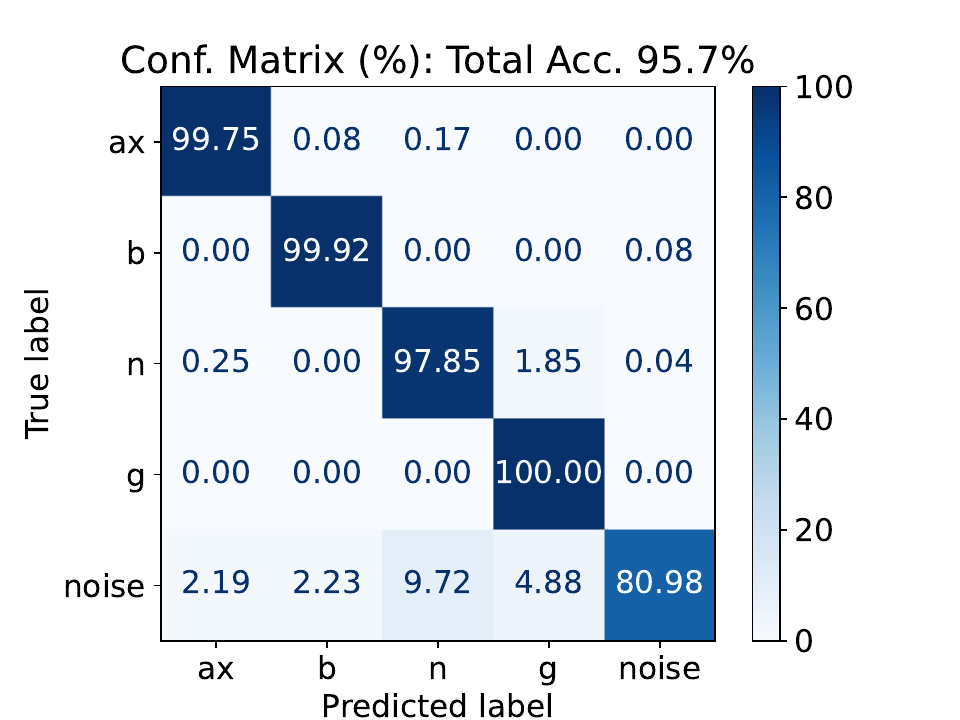}}
        \caption{RM\_B\_2}
        \label{fig:cv-RM_572C_2}
     \end{subfigure}
     ~
     \begin{subfigure}[b]{0.15\textwidth}
        \centering
        \hbox{\hspace{0em} \includegraphics[width=3.5cm]{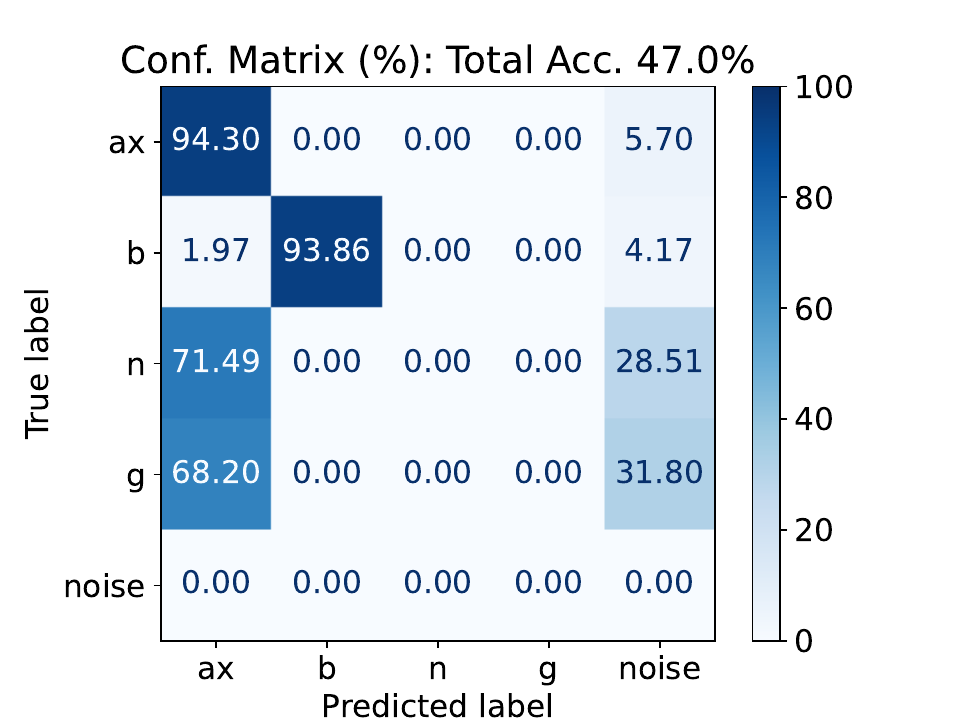}}
        \caption{RM\_C\_1}
        \label{fig:cv-RM_573C_1}
     \end{subfigure}
     ~
     \begin{subfigure}[b]{0.15\textwidth}
        \centering
        \hbox{\hspace{0em} \includegraphics[width=3.5cm]{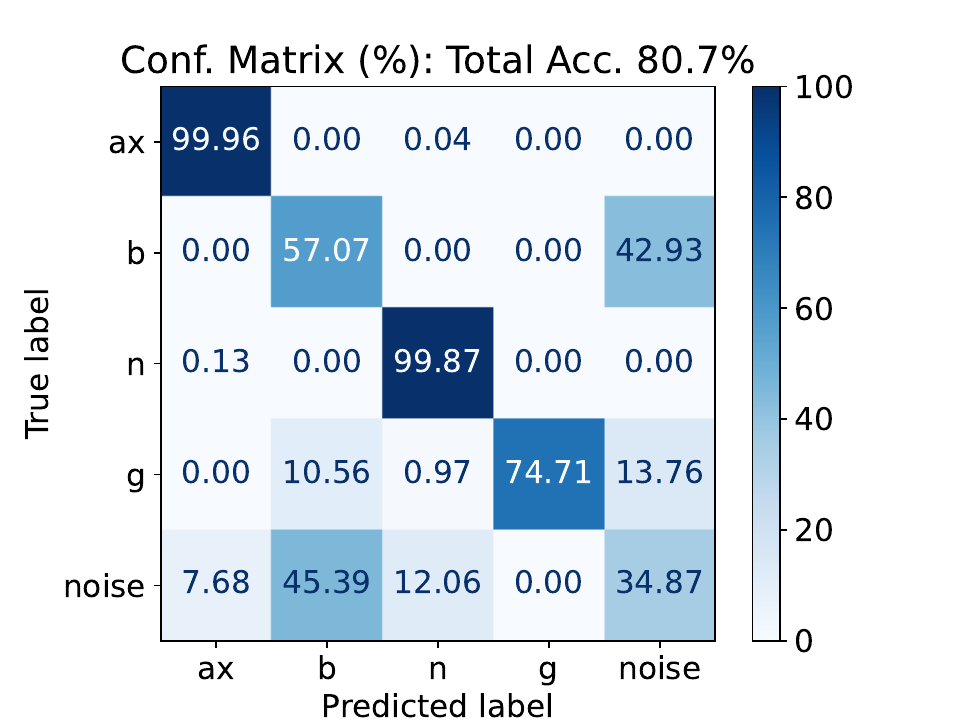}}
        \caption{RM\_C\_2}
        \label{fig:cv-RM_573C_2}
     \end{subfigure}
     ~
     
    \caption{ Cross-validation using LG classifier over the different OTA data collection instances. Each plot indicates what dataset is missing from the training set and the resulting model performance on that specific dataset at inference time. }
    \label{fig:cross-validation-OTA}
\end{figure}

\noindent $\bullet$ \textbf{Single-protocol results.}
 Fig. \ref{fig:LG-SM-avg-OTA} shows that both LG and SM models achieve an average classification accuracy $\ge 99\%$ on \emph{time-split} test data, with SM architecture slightly out-performing the LG one, possibly due to  \textit{overfitting} for the LG model.  Intuitively, a smaller variation of input patterns may be easier to \textit{learn} compared to randomly simulated scenarios in the synthetic dataset. 
Fig.~\ref{fig:cross-validation-OTA} shows the effect of the cross-validation on \emph{scenario-split} data. This highlights that RM\_C constitutes a more unique and challenging environment for our protocol classifier compared to other locations. %This highlights the need to expose our model to a wide variety of wireless scenarios to enable the model to generalize well. 

%In order to assess how different channel conditions affect the signal features learned by our classifier,

%Hence, in order to train our classifier to be as general as possible, we need to expose our model to a wide variety of wireless scenarios.

\begin{figure}
     \centering
     \begin{subfigure}[b]{0.22\textwidth}
        \centering
        \hbox{\hspace{0em} \includegraphics[width=4cm]{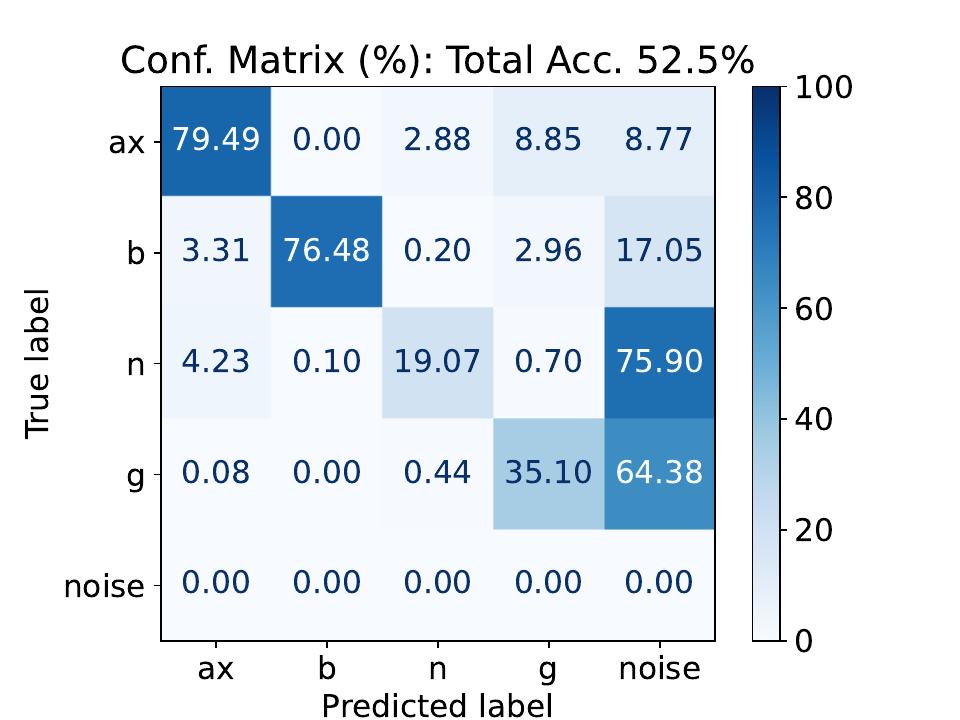}}
        \caption{W/O Power norm. }
        \label{fig:noRMS-0}
     \end{subfigure}
     ~
     \begin{subfigure}[b]{0.22\textwidth}
        \centering
        \hbox{\hspace{0em} \includegraphics[width=4cm]{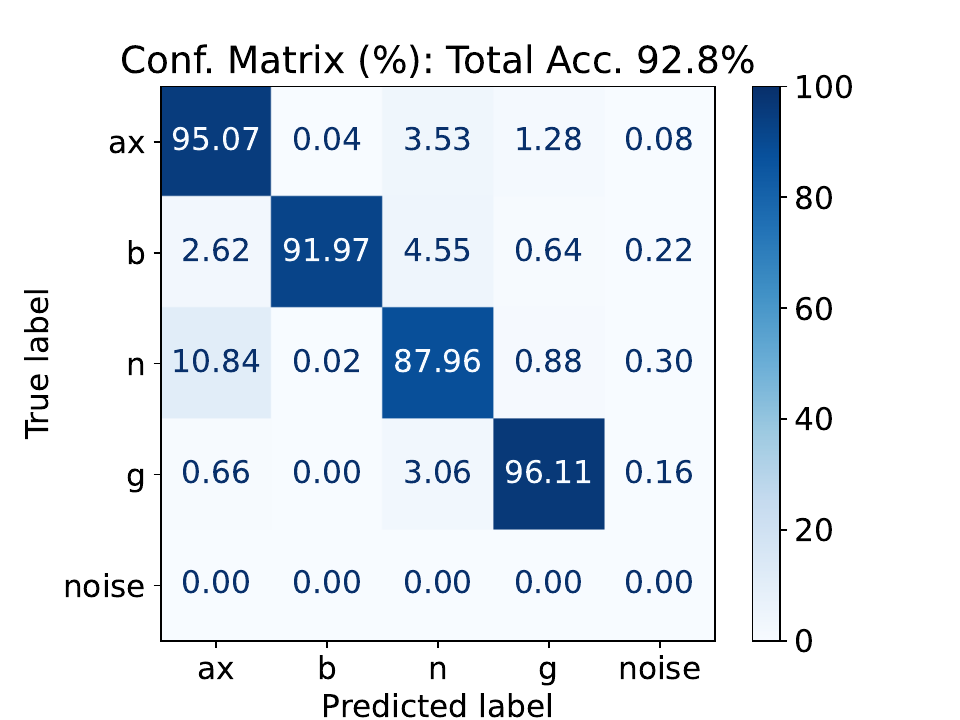}}
        \caption{W/ Power norm.}
        \label{fig:yesRMS-0}
     \end{subfigure}
     
        \caption{Impact of power normalization for low SNR conditions, i.e. $P_{\text{Tx}} = 0$ dBm, on test data collected in RM\_C. }
        \label{fig:RMSimpact}
\end{figure}

\noindent $\bullet$ \textbf{Overlapping-protocol results.}
Table \ref{tab:OTA_overlap_avg_accuracy} shows the average test accuracy on the complete OTA test dataset and, separately, a detailed view of accuracy for each individual overlapping transmission dataset. We use three different metrics to determine the performance: 1) percent of ``exact" classifications, i.e. correctly classifying both protocols present in the received signals, 2) percent of times we correctly recognize a ``single" protocol and 3) the Area-Under-the-Curve (AUC) of the trained classifier. %Looking at results for each overlapping configuration, we see that test samples for RM\_537\_O2, i.e. the case in which a larger portion of the spectrum is observed, lead to higher ``exact" accuracy among the two overlapping configurations. 
These results show that
%, while observing the interfering signals only in the time-domain, 
our model tested on RM\_C\_O2 is still able to correctly detect at least one protocol $94.4\%$ and $93.1\%$ of times for $25\%$ and $50\%$ cases, respectively, and even able to detect both signals at the same time with $75.7\%$ and $66\%$ of accuracy for $R = 25\%$ and $R = 50\%$, respectively, which is not possible using legacy wireless hardware. 
%On the other hand, when the transceivers are tuned on the same channel and bandwidth, i.e. RM\_C\_O1, the ``exact" accuracy results slightly lowered, due to narrower bandwidth that doesn't allow observations of signals over their whole occupied spectrum. Interestingly, while comparing overlapping configurations, we observe the trend is inverted when comparing one overlapping case to the other: while in RM\_C\_O2 we observe a lower ``exact" accuracy (i.e. $66\%$) with $R = 50\%$, due to wider range of frequency selective interference, the same property seem to boost ``exact" accuracy for the RM\_C\_O1 case (i.e. $72\%$), as interference become more prominent and more spectral information becomes available. 
Finally, we note that single-protocol accuracy is nearly perfect for RM\_C\_O1 (i.e. $99.9\%$ in both overlap configurations), which suggests that incumbent signals that occupy most of the observed spectrum are correctly classified the majority of times. 

\begin{table}[!t]
\resizebox{\columnwidth}{!}{
\begin{tabular}{l|rrrrr|}
\cline{2-6} 
& \multicolumn{5}{c|}{\textbf{Dataset}} \\
\cline{2-6} 
 & \multicolumn{1}{l|}{OTA (all)} & \multicolumn{2}{l|}{RM\_C\_O1} & \multicolumn{2}{l|}{RM\_C\_O2} \\ \cline{1-6} 
 \multicolumn{1}{|l|}{\textbf{Metric}} & \multicolumn{1}{l|}{} & \multicolumn{1}{l|}{$R = 25\%$} & \multicolumn{1}{l|}{$R = 50\%$} & \multicolumn{1}{l|}{$R = 25\%$} & \multicolumn{1}{l|}{$R = 50\%$} \\ \hline
\multicolumn{1}{|l|}{\% Exact (2/2)} & \multicolumn{1}{r|}{87.5} & 55.8 & \multicolumn{1}{r|}{72} & 75.7 & 66 \\ 
\multicolumn{1}{|l|}{\% Single (1/2)} & \multicolumn{1}{r|}{98.2} & 99.9 & \multicolumn{1}{r|}{99.9} & 94.4 & 93.1 \\ \cline{1-6}
\multicolumn{1}{|l|}{AUC} & \multicolumn{1}{r|}{0.98} & 0.878 & \multicolumn{1}{r|}{0.954} & 0.963 & 0.929 \\ \hline
\end{tabular}
}
\caption{Average test accuracy and AUC. Classifier is trained and tested over data covering all OTA locations, including single and overlap cases.}
    \label{tab:OTA_overlap_avg_accuracy}
\end{table}

Table \ref{tab:OTA_overlap_avg_protocol_acc} presents Precision\footnote{Precision = TP/(TP+FP); TP = True Positive; FP = False Positives.}, Recall\footnote{Recall = TP/(TP+FN); FN = False Negative.} % F1-score\footnote{f1-score = 2(Precision $\times$ Recall)/(Precision + Recall)}, 
and Support samples for each individual protocol in all receiver configurations. Precision in both RM\_C\_O1 cases shows higher scores for protocols 802.11b, 802.11g, and 802.11n, with a particular emphasis on 802.11b and 802.11g, as these protocols have more samples as incumbents than the others. Overall, the average Precision and Recall scores for all overlapping configurations and protocols are approximately $0.87$ and $0.86$, respectively. These results demonstrate the accuracy and completeness of positive predictions achieved by the proposed multi-protocol classifier architecture in this challenging interference scenario.

%To verify our intuition, Table \ref{tab:OTA_overlap_avg_protocol_acc} shows Precision\footnote{Precision = TP/(TP+FP); TP = True Positive; FP = False Positives.}, Recall\footnote{Recall = TP/(TP+FN); FN = False Negative.}, %f1-score\footnote{f1-score = 2(Precision $\times$ Recall)/(Precision + Recall)} 
%and number of Support samples for each individual protocol in all receiver configurations. If we look closely at Precision  in both RM\_C\_O1 cases, we notice that protocol 802.11b, 802.11g and 802.11n are the ones with higher score, with a particular bias toward 802.11b and 802.11g, i.e. the ones presenting the most number of samples as incumbents. More generally, we observe from Table \ref{tab:OTA_overlap_avg_protocol_acc} that Precision and Recall scores for all overlapping configurations and all protocols considered average to $0.87$ and $0.86$, respectively, which shows satisfying results in terms of accuracy and completeness of positive predictions for the proposed multi-protocol classifier architecture in this challenging interference scenario.

\begin{table}[t]
\resizebox{\columnwidth}{!}{
\begin{tabular}{|l|rrrrrr|}
\hline
Protocol & \multicolumn{2}{c|}{Precision} & \multicolumn{2}{c|}{Recall} & \multicolumn{2}{c|}{Support} \\ \hline
OTA (all) & \multicolumn{2}{l}{} & \multicolumn{2}{l}{} & \multicolumn{2}{l|}{} \\ \hline
802.11ax & \multicolumn{2}{c|}{0.94} & \multicolumn{2}{c|}{0.93} & \multicolumn{2}{c|}{25482} \\
802.11b & \multicolumn{2}{c|}{0.94} & \multicolumn{2}{c|}{0.98} & \multicolumn{2}{c|}{26154} \\
802.11n & \multicolumn{2}{c|}{0.9} & \multicolumn{2}{c|}{0.88} & \multicolumn{2}{c|}{26106} \\
802.11g & \multicolumn{2}{c|}{0.91} & \multicolumn{2}{c|}{0.91} & \multicolumn{2}{c|}{26106} \\ \hline
RM\_C\_O1 & \multicolumn{1}{l}{$R = 25\%$} & \multicolumn{1}{l|}{$R = 50\%$} & \multicolumn{1}{l}{$R = 25\%$} & \multicolumn{1}{l|}{$R = 50\%$} & \multicolumn{1}{l}{$R = 25\%$} & \multicolumn{1}{l|}{$R = 50\%$} \\ \hline
802.11ax & 0.72 & \multicolumn{1}{r|}{0.83} & 0.75 & \multicolumn{1}{r|}{0.86} & 2808 & 2808 \\
802.11b & 1 & \multicolumn{1}{r|}{1} & 1 & \multicolumn{1}{r|}{0.99} & 2808 & 2808 \\
802.11n & 0.77 & \multicolumn{1}{r|}{0.85} & 0.62 & \multicolumn{1}{r|}{0.83} & 2808 & 2808 \\
802.11g & 0.86 & \multicolumn{1}{r|}{0.96} & 0.78 & \multicolumn{1}{r|}{0.83} & 2808 & 2808 \\ \hline
RM\_C\_O2 & \multicolumn{1}{l}{$R = 25\%$} & \multicolumn{1}{l|}{$R = 50\%$} & \multicolumn{1}{l}{$R = 25\%$} & \multicolumn{1}{l|}{$R = 50\%$} & \multicolumn{1}{l}{$R = 25\%$} & \multicolumn{1}{l|}{$R = 50\%$} \\ \hline
802.11ax & 0.98 & \multicolumn{1}{r|}{0.97} & 0.97 & \multicolumn{1}{r|}{0.9} & 4329 & 4329 \\
802.11b & 0.82 & \multicolumn{1}{r|}{0.87} & 0.93 & \multicolumn{1}{r|}{0.96} & 4329 & 4329 \\
802.11n & 0.91 & \multicolumn{1}{r|}{0.78} & 0.8 & \multicolumn{1}{r|}{0.86} & 4329 & 4329 \\
802.11g & 0.86 & \multicolumn{1}{r|}{0.74} & 0.92 & \multicolumn{1}{r|}{0.8} & 4329 & 4329 \\ \hline
\end{tabular}
}
\caption{Test accuracy for individual protocols in single and overlapping transmissions. Classifier is trained and tested over data covering all OTA locations, including single and overlap cases.}
    \label{tab:OTA_overlap_avg_protocol_acc}

\end{table}

\noindent $\bullet$ \textbf{Impact of power normalization.}
We also study the impact of power normalization on an additional set of data collected under different transmission powers. Specifically, for this experiment we train our model on data collected from all locations (see Table \ref{tab:OTAdata}) but we test it on a completely different set of samples collected in RM\_C with transmissions performed at power $P_{\text{Tx}} = \{30, 20, 10, 0\}$ dBm respectively. Table \ref{tab:RMSimpact-accuracy} shows the average accuracy for each of the power profiles when power normalization is applied. While the average accuracy is consistently improved for all protocols, the most significant boost in performance is on the lower SNR end, reaching a staggering $40.3\%$ performance improvement at $P_{\text{Tx}} = 0$ dBm without the need to include any additional power variation samples in the dataset. 

\begin{table}[!t]
    \centering
    \begin{tabular}{|l| c c |}
        \hline
        $P_{\text{Tx}}$ (dBm) & \textbf{W/O Power Norm.} & \textbf{W/ Power Norm.} \\
        \hline
        0 & 52.5 & 92.8 \\
        10 & 93.8 & 95.9 \\
        20 & 93.9 & 95.3 \\
        30 & 97.3 & 97.6 \\
        \hline
    \end{tabular}
    \caption{Average test accuracy (\%) on multiple power profiles with and without signal power normalization.}
    \label{tab:RMSimpact-accuracy}
\end{table}

\subsubsection{Real-Time Latency and Power Results}

Building a pipeline with modular processing blocks that can operate in parallel is one of the keys to deploying a real-time classification system on an edge device. For example, as seen in Table \ref{tab: pipeline performance}, when using TensorRT, the Signal processing and ML inference blocks take nearly the same amount of time to process a sample. However, our pipeline allows for these processes to run in parallel producing a prediction every $10.9$ \unit{\milli\second} - equivalent to $92$ predictions per second.  
Table \ref{tab: pipeline performance} shows the power consumption and time to complete each block for all power profiles except 4-MAXP CORE DENVER which failed to run our pipeline. The top performance and most efficient power profiles from our testing are in bold. There is a clear trade off between higher performance in terms of predictions per second and higher power usage. For all power profiles, TensorRT provides better performance and is more efficient.

\begin{table}
\centering
\resizebox{\linewidth}{!}{%
\begin{tabular}{|l|c|c|c|c|c|} 
\hline
 & 0-MAXN & 1-MAXQ & 2-MAXP & 3-MAXP & 5-MAXN \\
 & ALL &  & ALL & ARM & ARM \\ 
\hline
\multicolumn{6}{|c|}{{\cellcolor[rgb]{0.753,0.753,0.753}}PyTorch} \\
\multicolumn{6}{|c|}{{\cellcolor[rgb]{0.753,0.753,0.753}}Average Power Consumption (\unit{\milli\watt})} \\ 
\hline
\rowcolor[rgb]{0.753,0.753,0.753} CPU & 1551 & 704 & 807 & 1457 & 1472 \\
\rowcolor[rgb]{0.753,0.753,0.753} DDR & 1595 & 842 & 900 & 999 & 1007 \\
\rowcolor[rgb]{0.753,0.753,0.753} GPU & 1074 & 486 & 366 & 434 & 435 \\
\rowcolor[rgb]{0.753,0.753,0.753} SOC & 1000 & 625 & 625 & 627 & 628 \\
\rowcolor[rgb]{0.753,0.753,0.753} ALL & 6933 & 4283 & 4338 & 5195 & 5242 \\ 
\hline
\multicolumn{6}{|c|}{{\cellcolor[rgb]{0.753,0.753,0.753}}Average Latency Time (\unit{\milli\second})} \\ 
\hline
\rowcolor[rgb]{0.753,0.753,0.753} Reciever & 0.4105 & 0.7252 & 0.44307 & \textbf{0.4103} & 0.4105 \\
\rowcolor[rgb]{0.753,0.753,0.753} Signal Pro. & 9.843 & 15.39 & 14.68 & 10.49 & 10.46 \\
\rowcolor[rgb]{0.753,0.753,0.753} ML Thread & 18.98 & 29.18 & 29.58 & 22.28 & 22.27 \\ 
\hhline{|======|}
\multicolumn{6}{|c|}{\textit{\textbf{TensorRT}}} \\
\multicolumn{6}{|c|}{Average Power Consumption (\unit{\milli\watt})} \\ 
\hline
CPU & 1579 & \textit{672} & 825 & 1515 & 1540 \\
DDR & 904 & \textit{780} & 958 & 1131 & 997 \\
GPU & 404 & \textit{264} & 436 & 629 & 578 \\
SOC & 610 & \textit{598} & 617 & 668 & 628 \\
ALL & 5170 & \textit{3923} & 4505 & 5668 & 5426 \\ 
\hline
\multicolumn{6}{|c|}{Average Latency Time (\unit{\milli\second})} \\ 
\hline
Reciever & 0.5095 & 0.6162 & 0.524 & 0.4797 & 0.5054 \\
Signal Pro. & 10.05 & 15.45 & 13.79 & \textbf{9.308} & 9.62 \\
ML Thread & 10.91 & 15.91 & 14.82 & 12.46 & \textbf{10.86} \\
\hline
ML inference$^*$ & 1.96 & 2.51 & 2.09 & 2.11 &  2.12 \\
\hline
\end{tabular}
}
\caption{T-PRIME LG pipeline performance for each block showing the \textbf{lowest latency} and \emph{lowest power} using different power profiles and ML inference optimization. The last line reports average inference times of T-PRIME LG for individual inference operations within the ML Thread. }
\label{tab: pipeline performance}
\end{table}

%We wrote a classification pipeline in Python to achieve the goal of real time protocol classification illustrated in Fig. \ref{Fig: system overview}. This pipeline consists of three primary functions that can run in parallel: receiver, signal processing, and ML inference. There are two thread safe FIFO queues, \texttt{q1} between the receiver and signal processing threads and \texttt{q2} between the signal processing and ML inference threads. This pipeline accepts several flags at run time to adjust several parameters including: center frequency, which ML model to use, and if various optimization options should be enabled. Building each of these functions as separate threads provides two main advantages. First, it allows us to parallelize our flow, removing concurrent dependencies and reducing the delay between making new predictions. Second, it allows us to modify or improve specific functions individually without having to make any changes to the other functions.

% \begin{figure}[ht]
%     \centering
%     \includegraphics[width=0.9\linewidth]{main_python.png}
%     \caption{Pipeline diagram.  {\color{red} TODO: separate this in 2 figs: 1) pipeline goes in architecture overview, 2) UI goes in separate demo figure}}
%     \label{Fig: pipeline}
% \end{figure}

\noindent $\bullet$ \textbf{Receiver thread:} The data buffer in the receiver thread presented a crucial I/O constraint that required careful engineering. A large buffer ensured enough samples for our model, but caused frequent buffer overflows, adding over 1s to reset the connection. On the other hand, a small buffer could lead to insufficient data for the ML model, requiring multiple reads of the buffer. Thus, we designed the buffer to hold a specific number of complex samples based on the ML model used. For the LG Transformer, we configured the buffer to capture $12,900$ complex samples.

%We found the data buffer in the receiver thread to be a key I/O constraint that required careful engineering. If we made the buffer very large, we are guaranteed enough samples for our model. However, do to I/O constraints this leads to frequent buffer overflow at between the SDR and the Jetson SOM. These buffer overflow can add over 1s to the timeline to reset the connection between the SDR and SOM. On the other hand, if we set the reciever data buffer too small, we may not have enough samples to feed to the ML model, resulting in multiple reads of the buffer before we have sufficient data. Therefore, we designed the buffer to hold a specific number of complex samples based on the ML model used. For the large Transformer, we configure the buffer to capture 12,900 complex samples.

\noindent $\bullet$ \textbf{Signal processing thread:} Understanding the effects of the low pass filter and re-sampling is crucial for this block. To down-sample from $31.25$MHz to $20$MHz, we perform rational re-sampling with a factor of $\frac{16}{25}$. For our LG Transformer trained with a $20$MHz bandwidth, we need $8,192$ complex samples, so the input size should be $8,192\times \frac{25}{16} = 12,800$. However, in practice, the output of this thread is slightly less than the required 8,192 samples. After empirical testing, we found that providing 12,900 complex samples as input gives the correct output size. This highlights both the complexity of deployed systems and the necessity for variable length inputs to the ML model.

%The key for this block is understanding the effects of the low pass filter and re-sampling. To down-sample from 31.25MHz to 20MHz we perform rational re-sampling with a factor of $\frac{16}{25}$. For our LG Transformer trained with a 20MHz bandwidth we need 8,192 complex samples so the input size should be $8,192\times \frac{25}{16} = 12,800$. However, in practice the output of this thread is less than the needed 8,192 samples. Instead we empirically determined we need 12,900 complex samples as the input to this block to give the correct output size. This issue highlights both the complexity of deployed systems and the need for variable length inputs to the ML model.

\noindent $\bullet$ \textbf{ML inference thread:} The key optimization for this thread was moving from PyTorch to TensorRT. The vast performance gains in the ML thread for TensorRT model shown in Table \ref{tab: pipeline performance}, without any loss of accuracy, highlight the importance of using well designed functional blocks which enable optimization of one function without making any changes to the others. 

\noindent $\bullet$ \textbf{Queue length management for real-time inference:}The above pipeline offers efficient use of computing resources, flexibility in optimizing each function, and reduced latency. However, it also presents potential bottlenecks with I/O and buffer management. To maintain low end-to-end latency, optimizing the queue length between blocks is crucial.
Long queues guarantee input availability for each function, maximizing parallelization gains and prediction rate. However, it significantly increases latency from signal reception to prediction. On the other hand, very short queues may lead to idle time, reducing parallelization efficiency and prediction rate.
Our pipeline's progressive processing times allow us to set short queues without starving any processes. However, jitters in processing time, IO blocks, and hardware limitations occasionally impact processing times dramatically. We found that setting both \texttt{q1} and \texttt{q2} to a length of $2$ minimized end-to-end delay while ensuring no thread was starved.

\subsubsection{Real-Time Deployment Results}

%\begin{figure}[tb]
%    \centering
%    \includegraphics[width=.8\linewidth]{figures/live_test.pdf}
%    \caption{T-PRIRME real-time prediction results in two new environments never seen during training.}
%    \label{Fig: real time results}
%\end{figure}

\begin{figure}
     \centering
     \begin{subfigure}[b]{0.22\textwidth}
        \centering
        \hbox{\hspace{0em} \includegraphics[width=4.1cm]{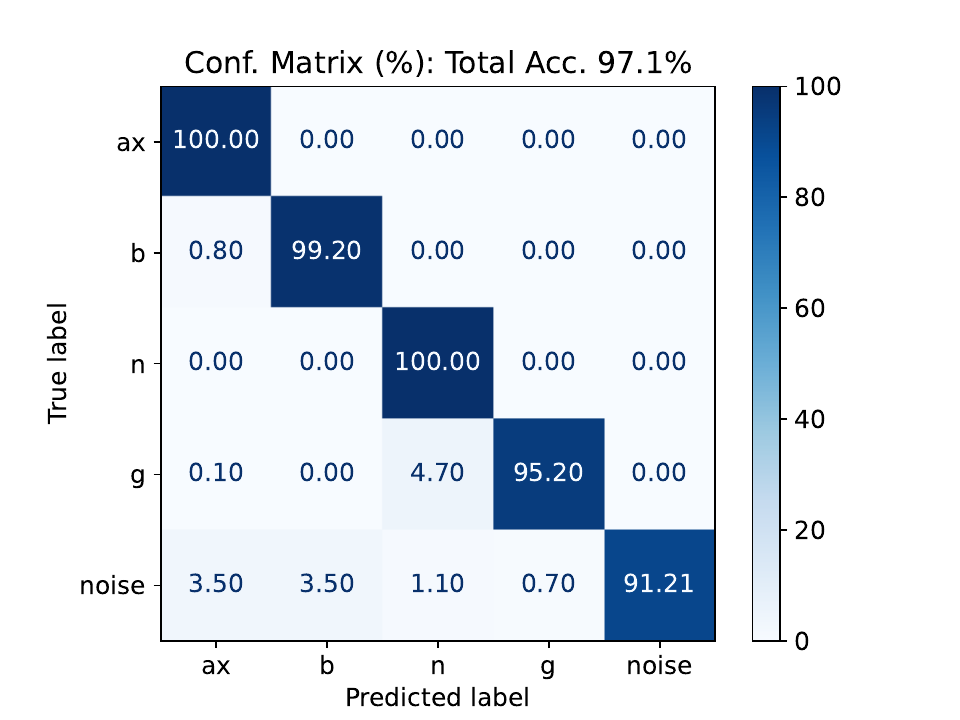}}
        \caption{RM\_D.}
        \label{fig:RM_D}
     \end{subfigure}
     ~
     \begin{subfigure}[b]{0.22\textwidth}
        \centering
        \hbox{\hspace{0em} \includegraphics[width=4.1cm]{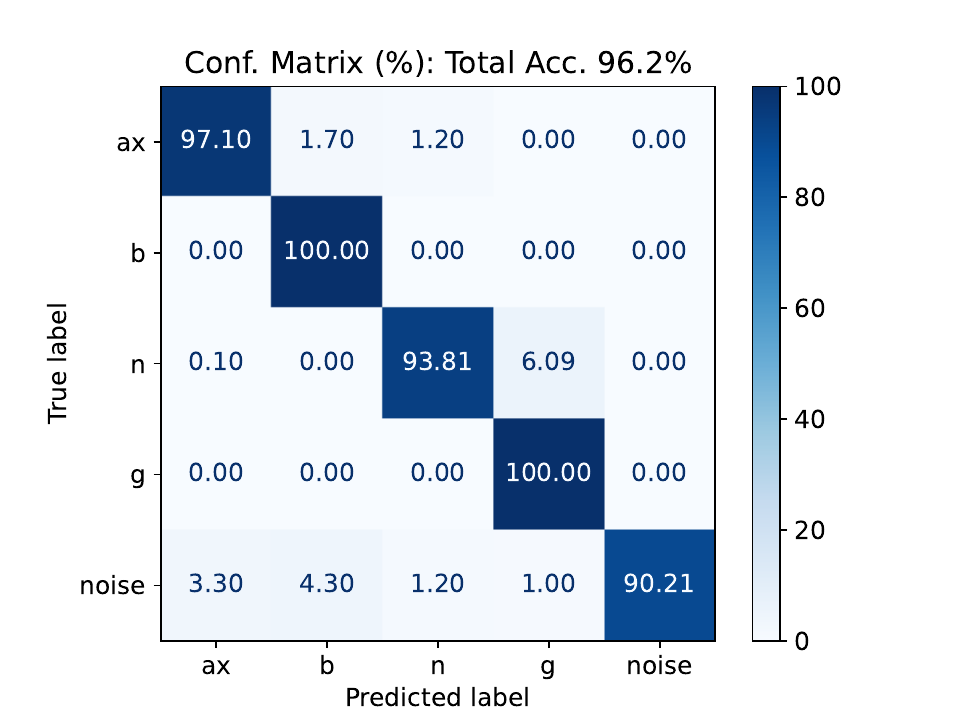}}
        \caption{RM\_E.}
        \label{fig:RM_E}
     \end{subfigure}
     
        \caption{T-PRIME real-time prediction results in two new environments, RM\_D and RM\_E, never seen during training.}
        \label{Fig: real time results}
\end{figure}

Fig. \ref{Fig: real time results} shows the confusion matrix for our Real-Time evaluation where we achieve an incredible $96.7$\% overall accuracy. From the confusion matrix, we see lowest accuracy is for the noise. However, given that other WiFi transmitters are operating in the area, some of the incorrect classifications for noise may in fact be correctly identifying other transmissions. Regardless, it is clear that T-PRIME has excellent accuracy in real-time predictions, even in environments it has never been exposed to during training.

\subsubsection{Discussion and Insights from Real-time Inference} Our implementation and testing of T-PRIME on the AIR-T device for real-time protocol classification have yielded valuable insights applicable to other AI/ML-enabled edge device classification tasks. Utilizing integrated SOM edge devices significantly reduces latencies and complexities often encountered in SDR-based systems. Parallelization plays a crucial role in increasing the prediction rate. Designing key functions as separate threads allows for more efficient resource utilization and ease of optimization for each function. Understanding I/O points and buffer management is essential for real-time systems.
%, although SOM devices generally mitigate these concerns. 
The ML Inference thread is the most resource-intensive, and optimizing this task offers significant performance gains. %in terms of latency and energy efficiency.

\section{Conclusion and future work}

T-PRIME achieves near-perfect classification accuracy for $4$ different WiFi protocols, outperforming legacy methods in challenging conditions. %This success demonstrates the capability of AI/ML-enabled edge devices for real-time WiFi protocol detection. 
T-PRIME proves superior to existing state of the art ML models at protocol detection, especially with low SNR and overlapping signals. While our current architecture employs fixed-length inputs, Transformers excel with variable length inputs, as shown by successful NLP models like ChatGPT. In the future, we will adapt our ML block to accommodate variable length inputs. %, optimizing for accuracy and prediction speed. 
As part of our real-time implementation, we identified different  bottlenecks and offered insights for solutions, which can benefit other AI/ML-based classification systems running on resource-constrained edge devices.  %Accurate protocol identification is a crucial step for effective spectrum sharing and cognitive radios, allowing intelligent systems to optimize transmission parameters. T-PRIME outlines the path towards achieving these goals.
Finally, we provide public access to our code and datasets at \cite{tprime_github}.

\section{Acknowledgment}
This work was funded by the Defence Science and Technology Laboratory (DSTL), a part of the UK Ministry of Defence.

\bibliographystyle{IEEEtran} % We choose the "plain" reference style
\bibliography{IEEEabrv,refs} % Entries are in the refs.bib file

\begin{thebibliography}{10}
\providecommand{\url}[1]{#1}
\csname url@samestyle\endcsname
\providecommand{\newblock}{\relax}
\providecommand{\bibinfo}[2]{#2}
\providecommand{\BIBentrySTDinterwordspacing}{\spaceskip=0pt\relax}
\providecommand{\BIBentryALTinterwordstretchfactor}{4}
\providecommand{\BIBentryALTinterwordspacing}{\spaceskip=\fontdimen2\font plus
\BIBentryALTinterwordstretchfactor\fontdimen3\font minus \fontdimen4\font\relax}
\providecommand{\BIBforeignlanguage}[2]{{%
\expandafter\ifx\csname l@#1\endcsname\relax
\typeout{** WARNING: IEEEtran.bst: No hyphenation pattern has been}%
\typeout{** loaded for the language `#1'. Using the pattern for}%
\typeout{** the default language instead.}%
\else
\language=\csname l@#1\endcsname
\fi
#2}}
\providecommand{\BIBdecl}{\relax}
\BIBdecl

\bibitem{zhang2022machine}
W.~Zhang and M.~Krunz, ``{Machine learning based protocol classification in unlicensed 5 GHz bands},'' in \emph{2022 IEEE International Conference on Communications Workshops (ICC Workshops)}.\hskip 1em plus 0.5em minus 0.4em\relax IEEE, 2022, pp. 752--757.

\bibitem{schmidt2017wireless}
M.~Schmidt, D.~Block, and U.~Meier, ``Wireless interference identification with convolutional neural networks,'' in \emph{2017 IEEE 15th international conference on industrial informatics (INDIN)}.\hskip 1em plus 0.5em minus 0.4em\relax IEEE, 2017, pp. 180--185.

\bibitem{jagannath2021multi}
A.~Jagannath and J.~Jagannath, ``Multi-task learning approach for automatic modulation and wireless signal classification,'' in \emph{ICC 2021-IEEE International Conference on Communications}.\hskip 1em plus 0.5em minus 0.4em\relax IEEE, 2021, pp. 1--7.

\bibitem{NIPS2017_3f5ee243}
A.~Vaswani, N.~Shazeer, N.~Parmar, J.~Uszkoreit, L.~Jones, A.~N. Gomez, {\L}.~Kaiser, and I.~Polosukhin, ``Attention is all you need,'' \emph{Advances in neural information processing systems}, vol.~30, 2017.

\bibitem{wolf-etal-2020-transformers}
T.~Wolf, L.~Debut, V.~Sanh, J.~Chaumond, C.~Delangue, A.~Moi, P.~Cistac, T.~Rault, R.~Louf, M.~Funtowicz \emph{et~al.}, ``Transformers: State-of-the-art natural language processing,'' in \emph{Proceedings of the 2020 conference on empirical methods in natural language processing: system demonstrations}, 2020, pp. 38--45.

\bibitem{9779340}
J.~Cai, F.~Gan, X.~Cao, and W.~Liu, ``Signal modulation classification based on the transformer network,'' \emph{IEEE Transactions on Cognitive Communications and Networking}, vol.~8, no.~3, pp. 1348--1357, 2022.

\bibitem{zheng2022fine}
Q.~Zheng, P.~Zhao, H.~Wang, A.~Elhanashi, and S.~Saponara, ``Fine-grained modulation classification using multi-scale radio transformer with dual-channel representation,'' \emph{IEEE Communications Letters}, vol.~26, no.~6, pp. 1298--1302, 2022.

\bibitem{hamidi2021mcformer}
S.~Hamidi-Rad and S.~Jain, ``Mcformer: A transformer based deep neural network for automatic modulation classification,'' in \emph{2021 IEEE Global Communications Conference (GLOBECOM)}.\hskip 1em plus 0.5em minus 0.4em\relax IEEE, 2021, pp. 1--6.

\bibitem{tprime_github}
{Genesys Lab}, ``{T-PRIME} {R}epository,'' \url{https://github.com/genesys-neu/t-prime}, 2023, [Accessed 2024-01-05].

\bibitem{9363693}
``{IEEE} {S}tandard for {I}nformation {T}echnology--{T}elecommunications and {I}nformation {E}xchange between systems - local and metropolitan area networks--specific requirements - part 11: Wireless lan medium access control ({MAC}) and physical layer ({PHY}) specifications,'' \emph{IEEE Std 802.11-2020 (Revision of IEEE Std 802.11-2016)}, pp. 1--4379, 2021.

\bibitem{terry2002ofdm}
J.~Terry and J.~Heiskala, \emph{OFDM wireless LANs: A theoretical and practical guide}.\hskip 1em plus 0.5em minus 0.4em\relax Sams publishing, 2002.

\bibitem{o2016convolutional}
T.~J. O’Shea, J.~Corgan, and T.~C. Clancy, ``Convolutional radio modulation recognition networks,'' in \emph{Engineering Applications of Neural Networks: 17th International Conference, EANN 2016, Aberdeen, UK, September 2-5, 2016, Proceedings 17}.\hskip 1em plus 0.5em minus 0.4em\relax Springer, 2016, pp. 213--226.

\bibitem{shi2019deep}
Y.~Shi, K.~Davaslioglu, Y.~E. Sagduyu, W.~C. Headley, M.~Fowler, and G.~Green, ``{Deep learning for RF signal classification in unknown and dynamic spectrum environments},'' in \emph{2019 IEEE International Symposium on Dynamic Spectrum Access Networks (DySPAN)}.\hskip 1em plus 0.5em minus 0.4em\relax IEEE, 2019, pp. 1--10.

\bibitem{elyousseph2021deep}
H.~Elyousseph and M.~L. Altamimi, ``Deep learning radio frequency signal classification with hybrid images,'' in \emph{2021 IEEE International Conference on Signal and Image Processing Applications (ICSIPA)}.\hskip 1em plus 0.5em minus 0.4em\relax IEEE, 2021, pp. 7--11.

\bibitem{8267032}
T.~J. O’Shea, T.~Roy, and T.~C. Clancy, ``Over-the-air deep learning based radio signal classification,'' \emph{IEEE Journal of Selected Topics in Signal Processing}, vol.~12, no.~1, pp. 168--179, 2018.

\bibitem{gravelle2019sdr}
C.~Gravelle and R.~Zhou, ``{SDR} demonstration of signal classification in real-time using deep learning,'' in \emph{2019 IEEE Globecom Workshops (GC Wkshps)}.\hskip 1em plus 0.5em minus 0.4em\relax IEEE, 2019, pp. 1--5.

\bibitem{huynh2020mcnet}
T.~Huynh-The, C.-H. Hua, Q.-V. Pham, and D.-S. Kim, ``{MCNet: An efficient CNN architecture for robust automatic modulation classification},'' \emph{IEEE Communications Letters}, vol.~24, no.~4, pp. 811--815, 2020.

\bibitem{zhang2023amc}
J.~Zhang, T.~Wang, Z.~Feng, and S.~Yang, ``{AMC-Net: An Effective Network for Automatic Modulation Classification},'' in \emph{ICASSP 2023-2023 IEEE International Conference on Acoustics, Speech and Signal Processing (ICASSP)}.\hskip 1em plus 0.5em minus 0.4em\relax IEEE, 2023, pp. 1--5.

\bibitem{8357902}
S.~Rajendran, W.~Meert, D.~Giustiniano, V.~Lenders, and S.~Pollin, ``Deep learning models for wireless signal classification with distributed low-cost spectrum sensors,'' \emph{IEEE Transactions on Cognitive Communications and Networking}, vol.~4, no.~3, pp. 433--445, 2018.

\bibitem{niu2021review}
Z.~Niu, G.~Zhong, and H.~Yu, ``A review on the attention mechanism of deep learning,'' \emph{Neurocomputing}, vol. 452, pp. 48--62, 2021.

\bibitem{dosovitskiy2020image}
A.~Dosovitskiy, L.~Beyer, A.~Kolesnikov, D.~Weissenborn, X.~Zhai, T.~Unterthiner, M.~Dehghani, M.~Minderer, G.~Heigold, S.~Gelly \emph{et~al.}, ``An image is worth 16x16 words: Transformers for image recognition at scale,'' \emph{arXiv preprint arXiv:2010.11929}, 2020.

\bibitem{wei2022emergent}
J.~Wei, Y.~Tay, R.~Bommasani, C.~Raffel, B.~Zoph, S.~Borgeaud, D.~Yogatama, M.~Bosma, D.~Zhou, D.~Metzler \emph{et~al.}, ``Emergent abilities of large language models,'' \emph{arXiv preprint arXiv:2206.07682}, 2022.

\bibitem{AIR-T}
\BIBentryALTinterwordspacing
``{A}{I}{R}-{T} {O}verview - {D}eepwave {D}igital {D}ocs --- docs.deepwavedigital.com,'' May 2022, [Accessed 30-07-2023]. [Online]. Available: \url{https://docs.deepwavedigital.com/AIR-T/}
\BIBentrySTDinterwordspacing

\bibitem{nvidiaQuickStart}
``{Q}uick {S}tart {G}uide : {N}{V}{I}{D}{I}{A} {D}eep {L}earning {T}ensor{R}{T} {D}ocumentation --- docs.nvidia.com,'' \url{https://docs.nvidia.com/deeplearning/tensorrt/quick-start-guide/index.html}, [Accessed 30-07-2023].

\end{thebibliography}
\vspace{12pt}
% \color{red}

\newpage

{\color{black}
%%% Yarkin edited %%%
\section*{Appendix}
\label{Sec:Appendix}

\subsection{Hyperparameter Exploration on Baseline Models}

In this section, baseline models are further explored to find optimal hyperparameters in order to compare with T-PRIME. In all tables, metrics are accuracy and cross-entropy loss respectively. In Tables \ref{tab:amcnet_all}, \ref{tab:resnet_all}, \ref{tab:1d_cnn_all}, and \ref{tab:mcformer_all}, we experimented with different values of slice length, batch size, and learning rate to find the optimal configuration over the synthetic dataset. In Table \ref{tab:overall_hyperparameters}, optimal hyperparameters for each baseline model is shown. Except ResNet, increasing slice length provided a better validation for the models. Additionally, decreasing the batch size allow the models train better, and smaller learning rates provide a better convergence. Afterwards, including T-PRIME LG and T-PRIME SM, all models are run for 50 epochs. Overall training and validation results including the fine-tuned baseline models and T-PRIME models are shown in Table \ref{tab:results_overall}, where \textit{M} is the sequence length, and \textit{S} is the slice length. The only difference between default and fine-tuned results over T-PRIME models are the number of epochs changing from 5 to 50, while the other (already optimized) hyperparameters remain same. We put the finalized comparisons that also include the runtimes in Table \ref{tab:comparison_models}.

\begin{table}[hbt]
\centering

\begin{tabular}{|c|c|c|c|c|}
\hline
\textbf{Slice Length} & \textbf{Batch Size} & \textbf{Learning Rate} & \textbf{Acc. (\%)} & \textbf{Loss} \\ \hline
128                   & 512                 & 1e-3                   & 68.8                        & 0.672                   \\
512                   & 512                 & 1e-3                   & 78.8                        & 0.464                   \\
1024                  & 512                 & 1e-3                   & 80.4                        & 0.438                   \\
2048                  & 512                 & 1e-3                   & 80.0                        & 0.432                   \\
4096                  & 512                 & 1e-3                   & 75.0                        & 0.706                   \\
8192                  & 32                  & 1e-3                   & 70.2                        & 0.713                   \\
8192                  & 64                  & 1e-3                   & 75.3                        & 0.568                   \\
8192                  & 128                 & 1e-3                   & 79.9                        & 0.450                   \\
8192                  & 256                 & 1e-3                   & 81.0                        & 0.433                   \\
8192                  & 512                 & 1e-4                   & 85.9                        & 0.324                   \\
\textbf{8192}         & \textbf{512}        & \textbf{2e-4}          & \textbf{86.2}               & \textbf{0.311}          \\
8192                  & 512                 & 5e-4                   & 86.4                        & 0.315                   \\
8192                  & 512                 & 8e-4                   & 84.0                        & 0.367                   \\
8192                  & 512                 & 1e-3                   & 82.6                        & 0.430                   \\
8192                  & 512                 & 5e-3                   & 51.4                        & 1.103                   \\
8192                  & 512                 & 1e-2                   & 30.0                        & 1.357                   \\ \hline
\end{tabular}
\caption{AMCNet Hyperparameter Exploration}
\label{tab:amcnet_all}
\end{table}

\begin{table}[hbt]
\centering

\begin{tabular}{|c|c|c|c|c|}
\hline
\textbf{Slice Length} & \textbf{Batch Size} & \textbf{Learning Rate} & \textbf{Acc. (\%)} & \textbf{Loss} \\ \hline
512                   & 512                 & 1e-3                   & 66.3                        & 0.770                   \\
1024                  & 32                  & 1e-3                   & 74.5                        & 0.561                   \\
1024                  & 64                  & 1e-3                   & 77.1                        & 0.517                   \\
1024                  & 128                 & 1e-4                   & 65.3                        & 0.786                   \\
1024                  & 128                 & 2e-4                   & 73.3                        & 0.580                   \\
1024                  & 128                 & 5e-4                   & 76.6                        & 0.516                   \\
\textbf{1024}         & \textbf{128}        & \textbf{1e-3}          & \textbf{78.3}               & \textbf{0.486}          \\
1024                  & 128                 & 5e-3                   & 29.2                        & 1.362                   \\
1024                  & 128                 & 1e-2                   & 27.8                        & 1.361                   \\
1024                  & 256                 & 1e-3                   & 74.6                        & 0.550                   \\
1024                  & 512                 & 1e-3                   & 71.2                        & 0.620                   \\
1024                  & 1024                & 1e-3                   & 68.2                        & 0.771                   \\
1024                  & 2048                & 1e-3                   & 64.2                        & 0.814                   \\
2048                  & 512                 & 1e-3                   & 65.1                        & 0.760                   \\
4096                  & 512                 & 1e-3                   & 62.8                        & 0.814                   \\
8192                  & 512                 & 1e-3                   & 67.4                        & 0.713                   \\
12288                 & 512                 & 1e-3                   & 48.2                        & 1.060                   \\
16384                 & 512                 & 1e-3                   & 34.9                        & 1.290                   \\ \hline       
\end{tabular}
\caption{ResNet Hyperparameter Exploration}
\label{tab:resnet_all}
\end{table}

\begin{table}[hbt]
\centering

\begin{tabular}{|c|c|c|c|c|}
\hline
\textbf{Slice Length} & \textbf{Batch Size} & \textbf{Learning Rate} & \textbf{Acc. (\%)} & \textbf{Loss} \\ \hline
512                   & 512                 & 1e-3                   & 45.6                        & 1.094                   \\
1024                  & 512                 & 1e-3                   & 52.3                        & 0.997                   \\
2048                  & 512                 & 1e-3                   & 39.3                        & 1.254                   \\
4096                  & 512                 & 1e-3                   & 44.6                        & 1.152                   \\
8192                  & 4                   & 1e-3                   & 55.8                        & 0.863                   \\
8192                  & 8                   & 1e-3                   & 57.2                        & 0.850                   \\
8192                  & 16                  & 1e-3                   & 57.6                        & 0.837                   \\
8192                  & 32                  & 1e-3                   & 57.6                        & 0.839                   \\
8192                  & 64                  & 1e-3                   & 57.8                        & 0.835                   \\
8192                  & 128                 & 1e-4                   & 60.8                        & 0.828                   \\
\textbf{8192}         & \textbf{128}        & \textbf{2e-4}          & \textbf{60.8}               & \textbf{0.815}          \\
8192                  & 128                 & 5e-4                   & 59.8                        & 0.832                   \\
8192                  & 128                 & 1e-3                   & 58.1                        & 0.835                   \\
8192                  & 128                 & 5e-3                   & 29.8                        & 1.345                   \\
8192                  & 128                 & 1e-2                   & 29.8                        & 1.345                   \\
8192                  & 256                 & 1e-3                   & 57.6                        & 0.864                   \\
8192                  & 512                 & 1e-3                   & 55.0                        & 0.891                   \\
8192                  & 1024                & 1e-3                   & 57.0                        & 0.942                   \\
8192                  & 2048                & 1e-3                   & 54.5                        & 1.057                   \\
12288                 & 512                 & 1e-3                   & 33.4                        & 1.287                   \\
16384                 & 512                 & 1e-3                   & 39.4                        & 1.305                   \\ \hline
\end{tabular}
\caption{1D CNN Hyperparameter Exploration}
\label{tab:1d_cnn_all}
\end{table}

\begin{table}[hbt]
\centering

\begin{tabular}{|c|c|c|c|c|}
\hline
\textbf{Slice Length} & \textbf{Batch Size} & \textbf{Learning Rate} & \textbf{Acc. (\%)} & \textbf{Loss} \\ \hline
1024                  & 16                  & 1e-3                   & 52.6                        & 0.964                   \\
1024                  & 32                  & 1e-3                   & 51.7                        & 0.968                   \\
1024                  & 64                  & 1e-3                   & 51.6                        & 0.970                   \\
1024                  & 128                 & 1e-3                   & 51.4                        & 0.974                   \\
1024                  & 256                 & 1e-3                   & 51.9                        & 0.959                   \\
2048                  & 16                  & 1e-3                   & 54.2                        & 0.903                   \\
2048                  & 32                  & 1e-3                   & 53.5                        & 0.903                   \\
2048                  & 64                  & 1e-4                   & 54.3                        & 0.909                   \\
\textbf{2048}         & \textbf{64}         & \textbf{2e-4}          & \textbf{55.2}               & \textbf{0.889}          \\
2048                  & 64                  & 5e-4                   & 54.7                        & 0.899                   \\
2048                  & 64                  & 1e-3                   & 54.9                        & 0.889                   \\
2048                  & 64                  & 5e-3                   & 28.3                        & 1.358                   \\
2048                  & 64                  & 1e-2                   & 28.3                        & 1.358                   \\
4096                  & 16                  & 1e-3                   & 53.7                        & 0.895                   \\ \hline
\end{tabular}
\caption{MCformer Hyperparameter Exploration}
\label{tab:mcformer_all}
\end{table}

\begin{table}[hb]
\centering
\resizebox{\columnwidth}{!}{
    \begin{tabular}{|c|cc|cc|cc|}
    \hline
    \multirow{2}{*}{\textbf{Models}} & \multicolumn{2}{c|}{\textbf{Slice Length}}                  & \multicolumn{2}{c|}{\textbf{Batch Size}}                    & \multicolumn{2}{c|}{\textbf{Learning Rate}}                 \\ \cline{2-7} 
                                     & \multicolumn{1}{c|}{\textbf{Default}} & \textbf{Fine-tuned} & \multicolumn{1}{c|}{\textbf{Default}} & \textbf{Fine-tuned} & \multicolumn{1}{c|}{\textbf{Default}} & \textbf{Fine-tuned} \\ \hline
    \textbf{AMCNet}                  & \multicolumn{1}{c|}{128}              & 8192                & \multicolumn{1}{c|}{512}              & 512                 & \multicolumn{1}{c|}{1e-3}             & 2e-4                \\
    \textbf{ResNet}                  & \multicolumn{1}{c|}{1024}             & 1024                & \multicolumn{1}{c|}{512}              & 128                 & \multicolumn{1}{c|}{1e-3}             & 1e-3                \\
    \textbf{1D CNN}                  & \multicolumn{1}{c|}{512}              & 8192                & \multicolumn{1}{c|}{512}              & 128                 & \multicolumn{1}{c|}{1e-3}             & 2e-4                \\
    \textbf{MCformer}                & \multicolumn{1}{c|}{128}              & 2048                & \multicolumn{1}{c|}{512}              & 64                  & \multicolumn{1}{c|}{1e-3}             & 2e-4                \\ \hline
    \end{tabular}
}
\caption{Default and Fine-tuned Hyperparameters of Baseline Models}
\label{tab:overall_hyperparameters}
\end{table}

\begin{table}[hb]
\centering

\resizebox{\columnwidth}{!}{
    \begin{tabular}{|c|cccc|cccc|}
    \hline
    \multirow{3}{*}{\textbf{Models}} & \multicolumn{4}{c|}{\textbf{Training}}                                                                                                 & \multicolumn{4}{c|}{\textbf{Validation}}                                                                                               \\ \cline{2-9} 
                                     & \multicolumn{2}{c|}{\textbf{Default}}                                        & \multicolumn{2}{c|}{\textbf{Fine-tuned}}                & \multicolumn{2}{c|}{\textbf{Default}}                                        & \multicolumn{2}{c|}{\textbf{Fine-tuned}}                \\ \cline{2-9} 
                                     & \multicolumn{1}{c|}{\textbf{Acc. (\%)}} & \multicolumn{1}{c|}{\textbf{Loss}} & \multicolumn{1}{c|}{\textbf{Acc. (\%)}} & \textbf{Loss} & \multicolumn{1}{c|}{\textbf{Acc. (\%)}} & \multicolumn{1}{c|}{\textbf{Loss}} & \multicolumn{1}{c|}{\textbf{Acc. (\%)}} & \textbf{Loss} \\ \hline
    \textbf{T-PRIME LG}              & \multicolumn{1}{c|}{92.4}               & \multicolumn{1}{c|}{0.213}         & \multicolumn{1}{c|}{96.4}               & 0.179         & \multicolumn{1}{c|}{93.5}               & \multicolumn{1}{c|}{0.166}         & \multicolumn{1}{c|}{97.5}               & 0.068         \\
    \textbf{T-PRIME SM}              & \multicolumn{1}{c|}{80.6}               & \multicolumn{1}{c|}{0.448}         & \multicolumn{1}{c|}{84.0}               & 0.405         & \multicolumn{1}{c|}{81.4}               & \multicolumn{1}{c|}{0.421}         & \multicolumn{1}{c|}{85.3}               & 0.334         \\
    \textbf{AMCNet}                  & \multicolumn{1}{c|}{68.6}               & \multicolumn{1}{c|}{0.653}         & \multicolumn{1}{c|}{93.4}               & 0.155         & \multicolumn{1}{c|}{68.8}               & \multicolumn{1}{c|}{0.672}         & \multicolumn{1}{c|}{93.5}               & 0.161         \\
    \textbf{ResNet}                  & \multicolumn{1}{c|}{64.7}               & \multicolumn{1}{c|}{0.667}         & \multicolumn{1}{c|}{79.9}               & 0.403         & \multicolumn{1}{c|}{66.3}               & \multicolumn{1}{c|}{0.770}         & \multicolumn{1}{c|}{80.3}               & 0.439         \\
    \textbf{1D CNN}                  & \multicolumn{1}{c|}{53.3}               & \multicolumn{1}{c|}{0.886}         & \multicolumn{1}{c|}{60.7}               & 0.700         & \multicolumn{1}{c|}{45.6}               & \multicolumn{1}{c|}{1.094}         & \multicolumn{1}{c|}{61.8}               & 0.757         \\
    \textbf{MCformer}                & \multicolumn{1}{c|}{48.7}               & \multicolumn{1}{c|}{1.034}         & \multicolumn{1}{c|}{56.6}               & 0.771         & \multicolumn{1}{c|}{50.1}               & \multicolumn{1}{c|}{0.992}         & \multicolumn{1}{c|}{57.1}               & 0.838         \\ \hline
    \end{tabular}
}
\caption{Overall Training and Validation Results of T-PRIME Models, and Default and Fine-tuned Baseline Models}
\label{tab:results_overall}
\end{table}

\begin{table}[hbt]
\centering

\resizebox{\columnwidth}{!}{
    \begin{tabular}{|c|cc|cc|cc|}
    \hline
    \multirow{2}{*}{\textbf{Models}} & \multicolumn{2}{c|}{\textbf{Input Size (M $\times$ S)}}            & \multicolumn{2}{c|}{\textbf{\# Parameters}}                 & \multicolumn{2}{c|}{\textbf{Runtime}}                       \\ \cline{2-7} 
                                     & \multicolumn{1}{c|}{\textbf{Default}} & \textbf{Fine-tuned} & \multicolumn{1}{c|}{\textbf{Default}} & \textbf{Fine-tuned} & \multicolumn{1}{c|}{\textbf{Default}} & \textbf{Fine-tuned} \\ \hline
    \textbf{T-PRIME LG}                  & \multicolumn{1}{c|}{64 $\times$ 128}          & 64 $\times$ 128            & \multicolumn{1}{c|}{6.8M}             & 6.8M                & \multicolumn{1}{c|}{0d 0h 28m}       & 0d 3h 43m        \\
    \textbf{T-PRIME SM}                  & \multicolumn{1}{c|}{24 $\times$ 64}          & 24 $\times$ 64            & \multicolumn{1}{c|}{1.6M}             & 1.6M                & \multicolumn{1}{c|}{0d 2h 26m}       & 0d 19h 2m        \\
    \textbf{AMCNet}                  & \multicolumn{1}{c|}{1 $\times$ 128}          & 1 $\times$ 8192            & \multicolumn{1}{c|}{462K}             & 269M                & \multicolumn{1}{c|}{3d 20h 37m}       & 0d 12h 34m          \\
    \textbf{ResNet}                  & \multicolumn{1}{c|}{1 $\times$ 1024}         & 1 $\times$ 1024            & \multicolumn{1}{c|}{162K}             & 162K                & \multicolumn{1}{c|}{0d 7h 27m}        & 2d 2h 48m           \\
    \textbf{1D CNN}                  & \multicolumn{1}{c|}{1 $\times$ 512}          & 1 $\times$ 8192            & \multicolumn{1}{c|}{4.1M}             & 67M                 & \multicolumn{1}{c|}{0d 7h 25m}        & 0d 5h 37m           \\
    \textbf{MCformer}                & \multicolumn{1}{c|}{1 $\times$ 128}          & 1 $\times$ 2048            & \multicolumn{1}{c|}{78K}              & 72K                 & \multicolumn{1}{c|}{2d 9h 13m}                & 1d 11h 3m           \\ \hline
    \end{tabular}
}
\caption{Comparisons between T-PRIME Models, and Default and Fine-tuned Baseline Models}
\label{tab:comparison_models}
\end{table}

~
~

\newpage

\subsection{Test Results on Channel Models}

In this section, T-PRIME models and optimized baseline models are compared on each individual channel model and all channel models over different values of SNR. Fig. \ref{fig:tprime-competitors-finetuned-accuracy} illustrates that even after the hyperparameter exploration, proposed T-PRIME LG model with fewer parameters in comparison, still consistently outperforms other optimized baseline models, achieving $96.7\%$ accuracy at SNR = $-15$ dB. Despite being outperformed by AMCNet and 1D CNN on several SNR conditions, T-PRIME SM still performs well with very few parameters compared to huge amount of parameters of these baseline models.

\begin{figure}[hb]
     \centering
     \begin{subfigure}[b]{0.5\textwidth}
        \centering
        \hbox{\hspace{-0.1em} \includegraphics[width=8.5cm]{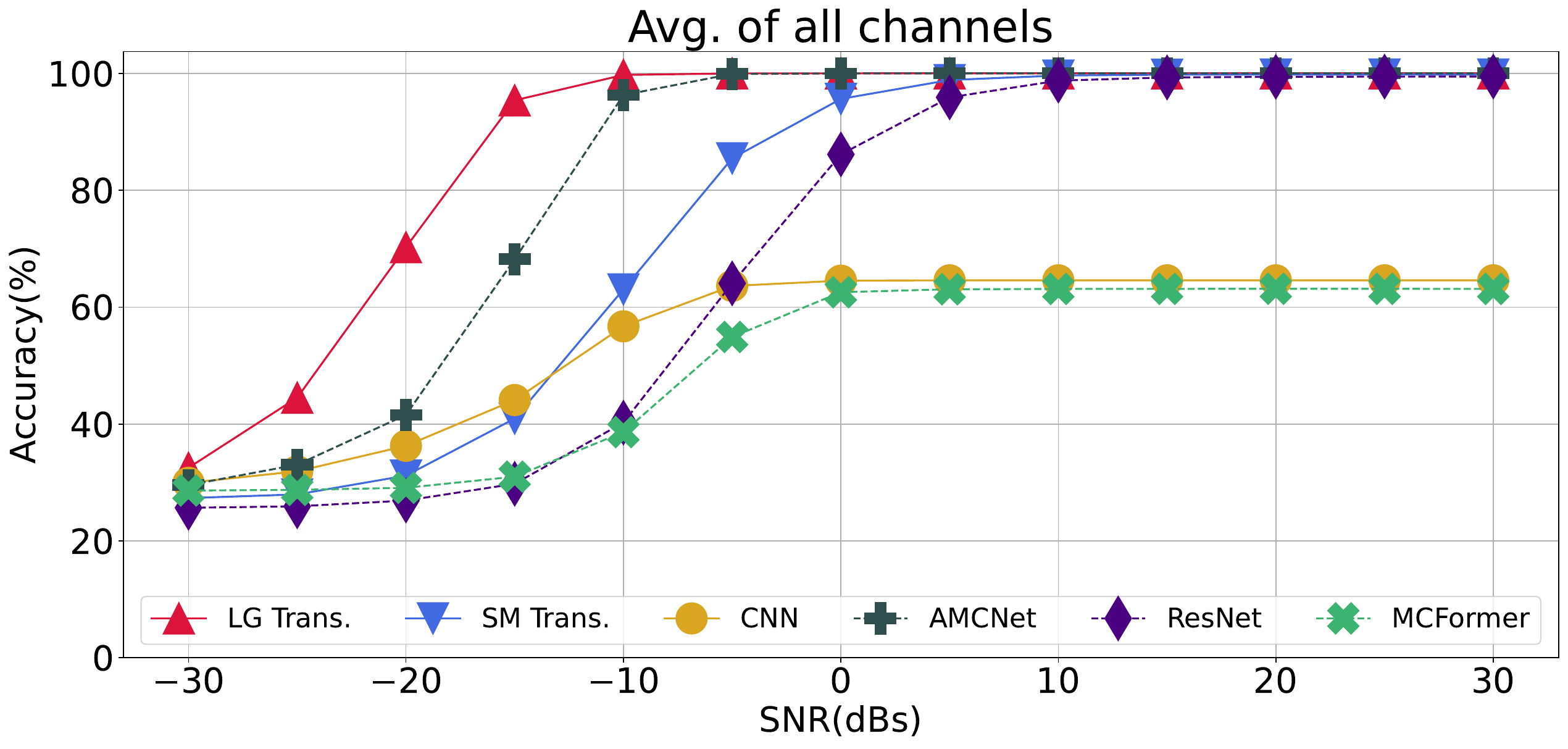}}

        \caption{}
        \label{fig:overall}
     \end{subfigure}
     \hfill
     \begin{subfigure}[b]{0.25\textwidth}
        \hbox{\hspace{-0.1em} \includegraphics[width=4.2cm]{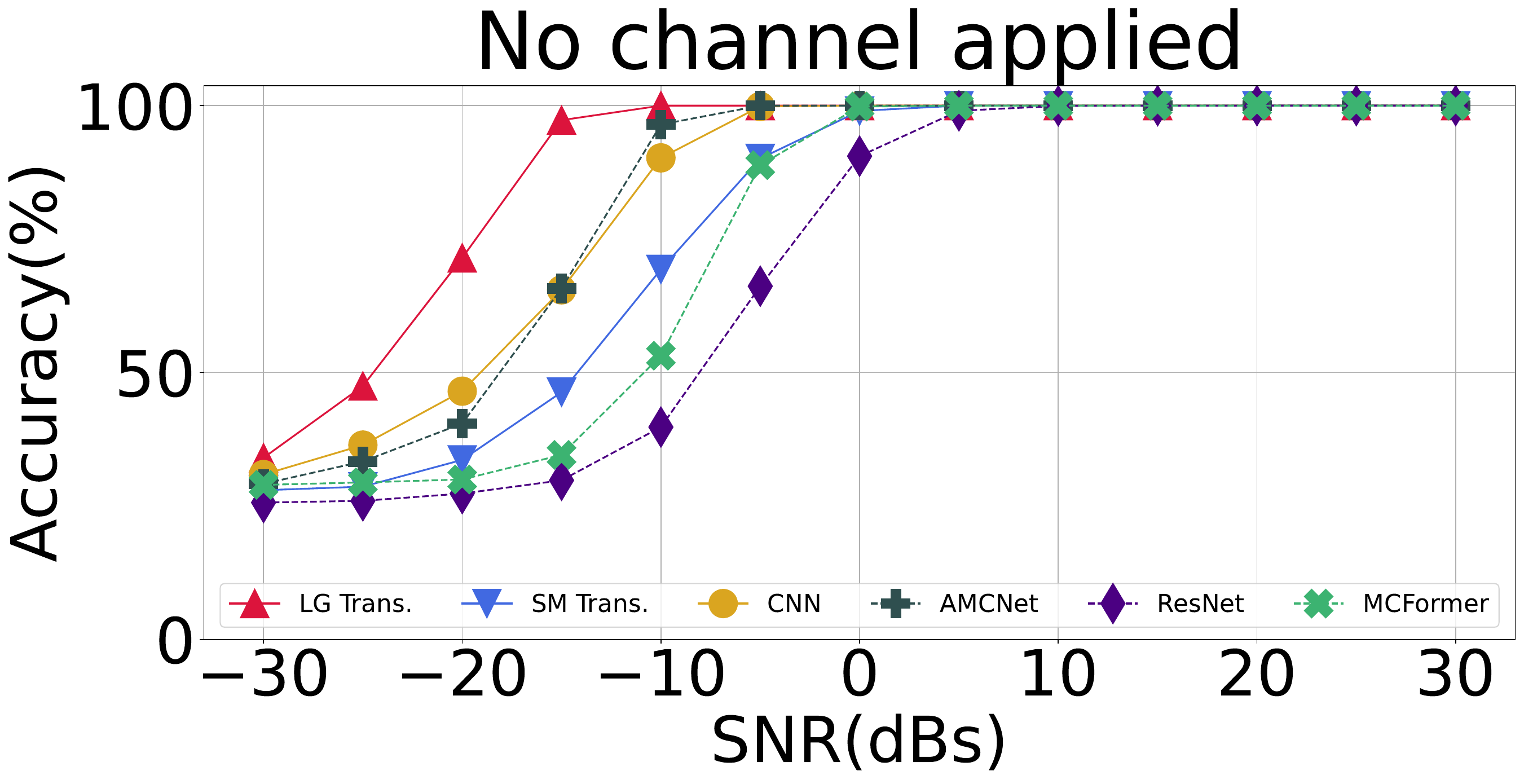}}
        \label{fig:none}
     \end{subfigure}
     \vspace{0.5em}
     \hfill
     \begin{subfigure}[b]{0.25\textwidth}
        \hbox{\hspace{-0.1em} \includegraphics[width=4.2cm]{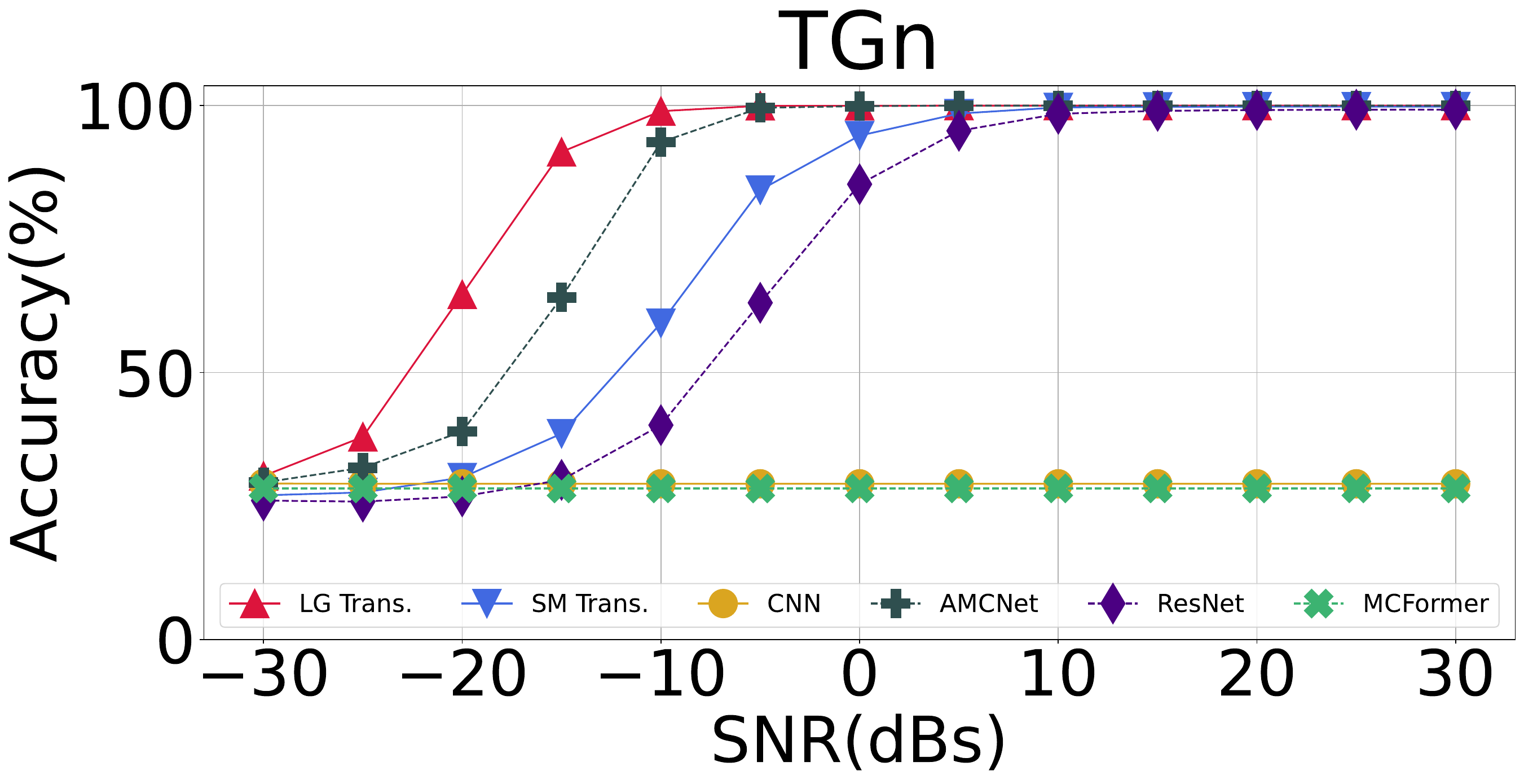}}
        \label{fig:rayleigh}
     \end{subfigure}
     \hfill
     \begin{subfigure}[b]{0.25\textwidth}
        \hbox{\hspace{-0.1em} \includegraphics[width=4.2cm]{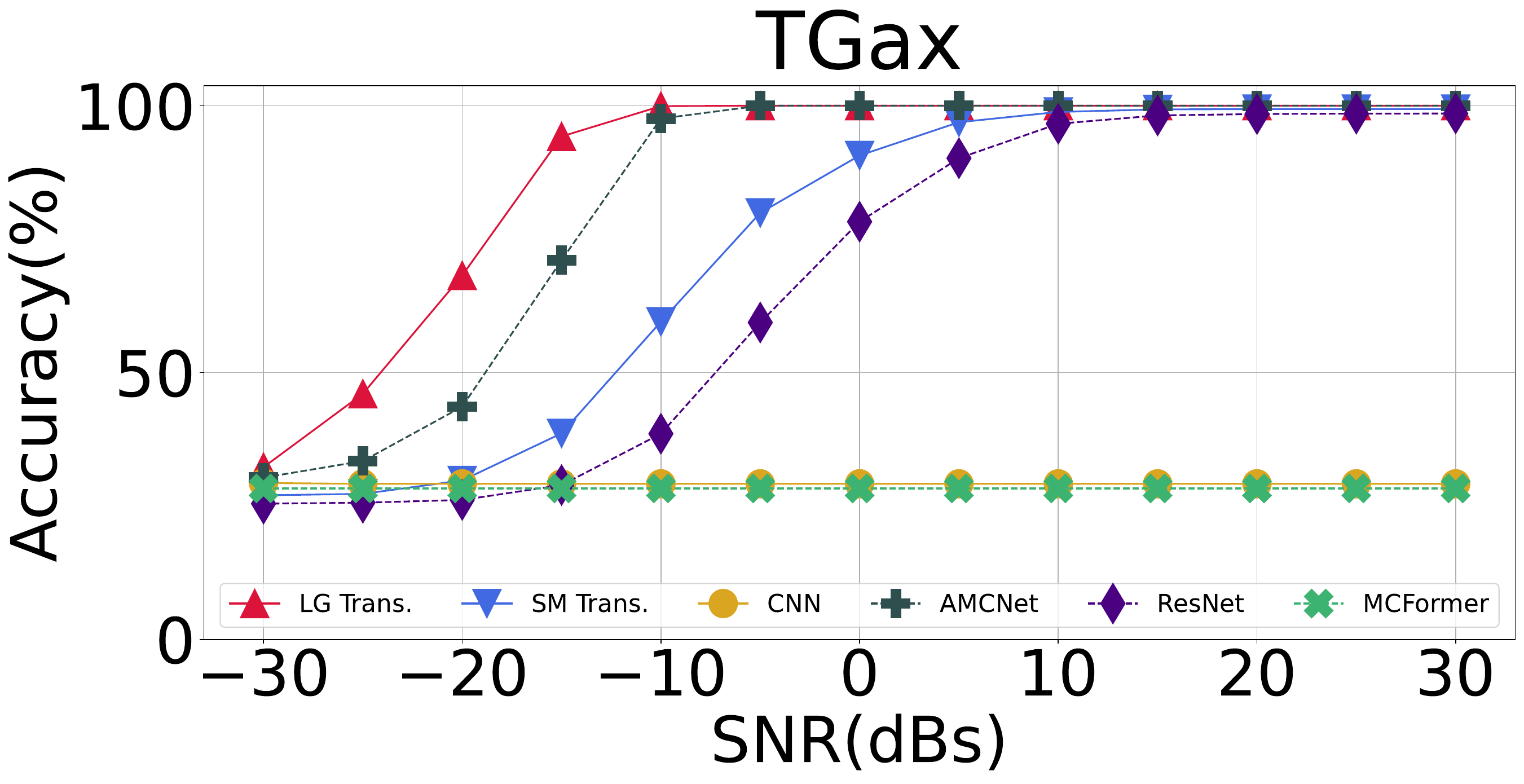}}
        \label{fig:tgax}
     \end{subfigure}
     \hfill
     \begin{subfigure}[b]{0.25\textwidth}
        \hbox{\hspace{-0.1em} \includegraphics[width=4.2cm]{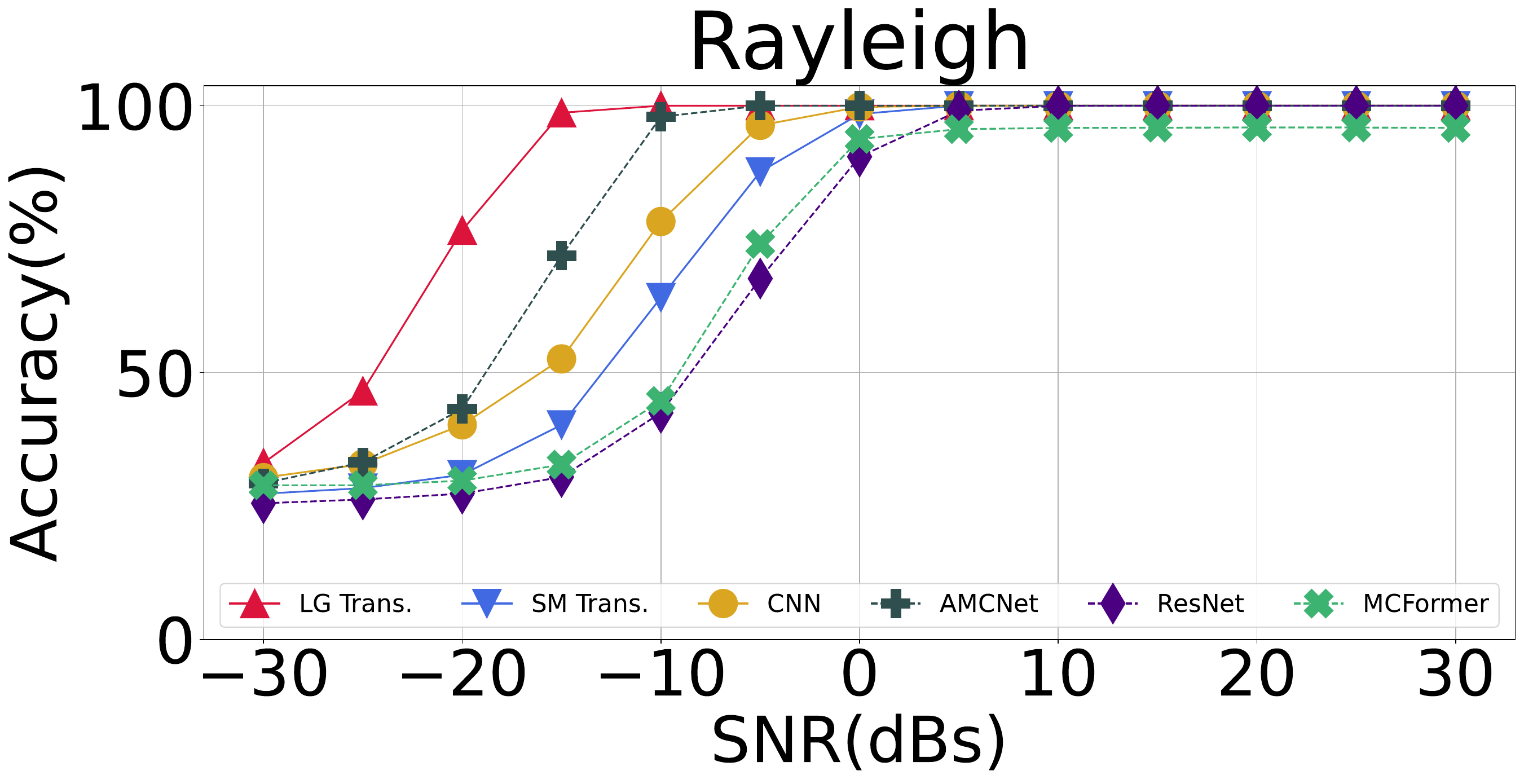}}
        \label{fig:tgn}
     \end{subfigure}

     \begin{subfigure}[b]{0.25\textwidth}
        \caption{}
     \end{subfigure}
     
        \caption{ (a) Comparison between SM, LG Transformer-based architectures and fine tuned baseline models tested on all channel models and different SNR conditions. The LG Transformer achieves the best overall accuracy and it retains high accuracy (i.e. $> 96.7\%$) for SNR as low as -15 dB. (b) Performance comparison for each individual channel model tested during simulation. }
        \label{fig:tprime-competitors-finetuned-accuracy}
\end{figure}

% Added by Mauro/Miquel
\subsection{Detailed accuracy metrics for multi-class classification on overlapping datasets}
%TODO: add description, make sure naming convention reflects names in the paper.

%In this final section, we present a breakdown of the classification accuracy of T-PRIME LG across all overlapping transmission configurations outlined in Table \ref{tab:wifi_combinations}. We introduce three distinct metrics for these results, one of which is novel with respect to the ones presented in Table \ref{tab:OTA_overlap_avg_accuracy}.

%We reiterate the exact and single accuracy metrics. Exact accuracy pertains to precisely identifying all and only the protocols being transmitted, while single accuracy denotes the percentage of transmissions in which we successfully detect at least one of the transmitted protocols, irrespective of any incorrect detections of non-transmitted protocols.

%The new metric, termed single restrictive accuracy, quantifies the instances in which we detect at least one of the transmitted protocols while simultaneously avoiding detection of any non-transmitted protocols. Specifically, this metric identifies as correct only those samples containing either one or both of the transmitted protocols, excluding any other sample types.

In this section, we present a detailed analysis of the classification accuracy of T-PRIME LG across various overlapping transmission configurations outlined in Table \ref{tab:wifi_combinations}. We introduce three distinct metrics for these results, one of which is innovative compared to the metrics presented in Table \ref{tab:OTA_overlap_avg_accuracy}.

We emphasize two specific accuracy metrics: Exact Accuracy and Single Accuracy. Exact Accuracy pertains to precisely identifying all and only the protocols being transmitted, while single accuracy denotes the percentage of transmissions where we successfully detect at least one of the transmitted protocols, regardless of any incorrect detection of protocols not present in the signal.

The new metric, referred to as Single Exact Accuracy, quantifies instances in which we detect at least one of the transmitted protocols while simultaneously avoiding the detection of any non-transmitted protocols. Specifically, this metric recognizes as correct only those samples containing either one or both of the transmitted protocols, counting a sample as incorrect if classifier outputs any other protocol combination.

\begin{table}[htbp]

\centering

\begin{tabular}{|l|c|c|c|c|c|c|c|}
\hline
& \multicolumn{6}{l|}{Overlapping configuration (incumbent - interferer)}  \\   
\hline
Metrics (\%) & \textbf{b - ax} & \textbf{b - g} & \textbf{b - n} & \textbf{n - ax} & \textbf{g - ax} & \textbf{g - n} \\
\hline
Exact Accuracy & 69.9 & 63.8 & 69.4 & 62.5 & 76.9 & 79.5 \\
Single Exact Accuracy & 80.2 & 76.5 & 79.4 & 70.7 & 82.2 & 82.6 \\
Single Accuracy & 99.8 & 100.0 & 99.9 & 99.9 & 100.0 & 100.0 \\
\hline
ax detected & \textbf{80.8} & 6.4 & 9.2 & \textbf{92.9} & \textbf{90.6} & 16.5 \\
b detected & \textbf{92.9} & \textbf{100.0} & \textbf{96.4} & 7.0 & 2.7 & 0.8 \\
n detected & 13.8 & 17.3 & \textbf{77.7} & \textbf{75.1} & 15.2 & \textbf{84.5} \\
g detected & 6.2 & \textbf{65.7} & 11.6 & 22.3 & \textbf{88.9} & \textbf{97.1} \\
noise detected & 0.0 & 0.0 & 0.0 & 0.0 & 0.0 & 0.0 \\
\hline

\end{tabular}

\caption{Global results}
\label{tab:exploded_OTA_global}

\bigskip

\begin{tabular}{|l|c|c|c|c|c|c|c|}
\hline
& \multicolumn{6}{l|}{Overlapping configuration (incumbent - interferer)}  \\   
\hline
Metrics (\%) & \textbf{b - ax} & \textbf{b - g} & \textbf{b - n} & \textbf{n - ax} & \textbf{g - ax} & \textbf{g - n} \\
\hline
Exact Accuracy & 72.6 & 97.9 & 90.9 & 38.9 & 84.8 & 89.1 \\
Single Exact Accuracy & 79.6 & 98.8 & 94.3 & 53.1 & 90.9 & 94.3 \\
Single Accuracy & 99.8 & 100.0 & 100.0 & 99.8 & 99.9 & 100.0 \\
\hline
ax detected & \textbf{92.4} & 0.2 & 1.5 & \textbf{97.8} & \textbf{99.8} & 4.0 \\
b detected & \textbf{84.7} & \textbf{100.0} & \textbf{95.4} & 19.7 & 7.6 & 1.7 \\
n detected & 16.5 & 1.0 & \textbf{96.2} & \textbf{45.9} & 1.5 & \textbf{98.2} \\
g detected & 4.0 & \textbf{98.0} & 4.2 & 27.2 & \textbf{86.2} & \textbf{92.7} \\
noise detected & 0.0 & 0.0 & 0.0 & 0.1 & 0.0 & 0.0 \\
\hline
\end{tabular}
\caption{Dataset RM\_C\_O1 (overlap ratio $R  = 25\%$)}
\label{tab:exploded_OTA_O1_25}
\bigskip

\begin{tabular}{|l|c|c|c|c|c|c|c|}
\hline
& \multicolumn{6}{l|}{Overlapping configuration (incumbent - interferer)}  \\   
\hline
Metrics (\%) & \textbf{b - ax} & \textbf{b - g} & \textbf{b - n} & \textbf{n - ax} & \textbf{g - ax} & \textbf{g - n} \\
\hline
Exact Accuracy & 82.8 & 58.8 & 70.5 & 40.3 & 69.6 & 96.3 \\
Single Exact Accuracy & 87.5 & 64.0 & 75.6 & 51.8 & 76.3 & 97.3 \\
Single Accuracy & 99.7 & 100.0 & 99.9 & 100.0 & 100.0 & 99.9 \\
\hline
ax detected & \textbf{93.1} & 1.4 & 5.1 & \textbf{79.8} &\textbf{ 95.9} & 1.6 \\
b detected & \textbf{94.0} & \textbf{100.0} & \textbf{93.3} & 2.3 & 1.2 & 1.1 \\
n detected & 12.3 & 34.7 & \textbf{85.9} & \textbf{74.2} & 22.5 & \textbf{98.8} \\
g detected & 0.3 & \textbf{62.6} & 19.3 & 45.9 & \textbf{78.4} & \textbf{98.4} \\
noise detected & 0.0 & 0.0 & 0.0 & 0.0 & 0.0 & 0.0 \\
\hline
\end{tabular}
\caption{Dataset RM\_C\_O1 (overlap ratio $R  = 50\%$)}
\label{tab:exploded_OTA_O1_50}
\bigskip

\begin{tabular}{|l|c|c|c|c|c|c|c|}
\hline
& \multicolumn{6}{l|}{Overlapping configuration (incumbent - interferer)}  \\   
\hline
Metrics (\%) & \textbf{b - ax} & \textbf{b - g} & \textbf{b - n} & \textbf{n - ax} & \textbf{g - ax} & \textbf{g - n} \\
\hline
Exact Accuracy & 49.5 & 33.4 & 28.1 & 97.6 & 71.4 & 55.0 \\
Single Exact Accuracy & 70.7 & 58.3 & 57.1 & 98.6 & 76.8 & 58.7 \\
Single Accuracy & 100.0 & 100.0 & 99.8 & 99.9 & 100.0 & 100.0 \\
\hline
ax detected & \textbf{53.3} & 20.8 & 26.7 & \textbf{99.1} & \textbf{72.6} & 41.3 \\
b detected & \textbf{99.7} & \textbf{100.0} &\textbf{ 99.7} & 1.2 & 0.0 & 0.0 \\
n detected & 10.4 & 21.8 & \textbf{31.3} & \textbf{98.5} & 23.2 & \textbf{57.7} \\
g detected & 19.2 & \textbf{35.0} & 17.2 & 0.3 & \textbf{99.7} & \textbf{99.5} \\
noise detected & 0.0 & 0.0 & 0.0 & 0.0 & 0.0 & 0.0 \\
\hline
\end{tabular}
\caption{Dataset RM\_C\_O2 (overlap ratio $R  = 25\%$)}
\label{tab:exploded_OTA_O2_25}
\bigskip

\begin{tabular}{|l|c|c|c|c|c|c|c|}
\hline
& \multicolumn{6}{l|}{Overlapping configuration (incumbent - interferer)}  \\   
\hline
Metrics (\%) & \textbf{b - ax} & \textbf{b - g} & \textbf{b - n} & \textbf{n - ax} & \textbf{g - ax} & \textbf{g - n} \\
\hline
Exact Accuracy & 64.6 & 49.4 & 75.6 & 97.6 & 81.5 & 63.2 \\
Single Exact Accuracy & 79.5 & 79.4 & 84.5 & 99.1 & 83.2 & 65.9 \\
Single Accuracy & 99.8 & 100.0 & 100.0 & 100.0 & 99.9 & 100.0 \\
\hline
ax detected & \textbf{71.4} & 9.3 & 10.1 & \textbf{99.5} & \textbf{86.4} & 34.1 \\
b detected & \textbf{96.9} & \textbf{100.0} & \textbf{99.5} & 0.4 & 0.0 & 0.0 \\
n detected & 15.2 & 11.3 & \textbf{83.0} & \textbf{98.2} & 16.8 & \textbf{68.3} \\
g detected & 5.6 & \textbf{51.4} & 5.3 & 0.4 & \textbf{98.7} & \textbf{99.7} \\
noise detected & 0.0 & 0.0 & 0.0 & 0.0 & 0.0 & 0.0 \\
\hline
\end{tabular}
\caption{Dataset RM\_C\_O2 (overlap ratio $R  = 50\%$)}
\label{tab:exploded_OTA_O2_50}
\end{table}

%A initial observation is that when two protocols are simultaneously transmitted, the model predominantly detects these protocols more effectively compared to the other one not being transmitted. This observation highlights the classifier's high accuracy in telling protocols apart between concurrently transmitted signals.

%Upon closer examination of the results, it becomes apparent that 802.11b is the protocol most readily identified by the model. This heightened detection rate could stem from several factors, including the distinctiveness of this protocol relative to others and the fact that we consistently transmit it as the incumbent signal.

%Another finding is that despite typically achieving superior detection accuracy for the incumbent signal compared to the interferer, Dataset RM\_C\_02 exhibits a deviation in this model's behavior for the last two overlap configurations (ax-g and n-g).

%Moreover, it is noteworthy to observe the model's struggle in identifying the 802.11g protocol when mixed with 802.11b. In this specific configuration, the former protocol presents the most significant challenge for the model in terms of identification. Nonetheless, the model continues to yield sufficiently satisfactory results despite this difficulty.

From Table \ref{tab:exploded_OTA_global} we observe an average of 86.88\% correct detection rates for protocol classes that are actually present within the overlapping signals in the test set, while detection rate for protocols not present is on average 7.16\%. This demonstrates how our classifier can accurately identify the protocols even in presence of active interference, and it is able to correctly determine both incumbent and interferer transmissions in the majority of cases.

Upon a more detailed analysis of the results, looking at Tables \ref{tab:exploded_OTA_O1_25}, \ref{tab:exploded_OTA_O1_50}, \ref{tab:exploded_OTA_O2_25} and \ref{tab:exploded_OTA_O2_50} it becomes evident that 802.11b is the protocol most easily identified by the model. This behavior may arise from a number of factors, including distinctiveness of its waveforms compared to others and oversampling applied to signals in this category in order to match other protocols' configurations in the rest of the dataset.

Another noteworthy observation is the distinct behavior of the model across the two datasets. In dataset RM\_C\_02, we generally observe superior detection accuracy for the incumbent signal compared to the interferer. However, in dataset RM\_C\_01, the model deviates from this pattern, exhibiting numerous instances where higher accuracy is achieved for the interferer.

Finally, it is noteworthy to observe the model's difficulty in identifying the 802.11g protocol particularly when mixed as an interferer with 802.11b. In this specific configuration, the former protocol presents the most significant challenge for the model in terms of identification. Nevertheless, the model overall continues to yield satisfactory results, motivating further research efforts in these overlapping scenarios.

}

% IEEE conference templates contain guidance text for composing and formatting conference papers. Please ensure that all template text is removed from your conference paper prior to submission to the conference. Failure to remove the template text from your paper may result in your paper not being published.

\end{document}